\documentclass{article}
\PassOptionsToPackage{numbers, compress}{natbib}
\usepackage[final]{neurips_2022}
\usepackage[utf8]{inputenc} 
\usepackage[T1]{fontenc}    
\usepackage{hyperref}       
\usepackage{url}            
\usepackage{booktabs}       
\usepackage{amsfonts}       
\usepackage{nicefrac}       
\usepackage{microtype} 
\usepackage{enumitem}
\usepackage{xcolor}
\usepackage{tablefootnote}
\usepackage{graphicx}
\usepackage{subfigure}

\usepackage{amssymb,amsmath,color,enumitem}
\usepackage[english]{babel}
\usepackage{tikz}
\usepackage{commath}
\usepackage{breqn}
\usepackage{natbib}
\usepackage{bbm}
\usepackage{bm}
\usepackage{algorithm}
\usepackage[noend]{algorithmic}
\usepackage{tablefootnote}
\usepackage{amsthm}

\newtheorem{cor}{Corollary}
\newtheorem{assum}{Assumption}
\newtheorem{lem}{Lemma}

\newtheorem{remark}{Remark}
\newtheorem{theorem}{Theorem}

\newtheorem{definition}{Definition}

\usepackage[symbol]{footmisc}

 \usepackage{threeparttable}
\usepackage{enumitem}

\newcommand{\R}{\mathbb R}

\newcommand{\rD}{\mathrm{D}}

\newcommand{\rI}{\mathrm{I}}

\newcommand{\rS}{\mathrm{S}}
\newcommand{\rT}{\mathrm{T}}

\newcommand{\bP}{\mathbbm{P}}
\newcommand{\bE}{\mathbbm{E}}
\newcommand{\vA}{\mathbf{A}}
\newcommand{\cA}{\mathcal A}

\newcommand{\cC}{\mathcal C}

\newcommand{\cE}{\mathcal E}
\newcommand{\cF}{\mathcal F}

\newcommand{\cL}{\mathcal L}

\newcommand{\cN}{\mathcal N}
\newcommand{\cO}{\mathcal O}

\newcommand{\cQ}{\mathcal Q}

\newcommand{\cS}{\mathcal S}
\newcommand{\cT}{\mathcal T}

\newcommand{\cV}{\mathcal V}

\newcommand{\bI}{\mathbf{I}}
\newcommand{\vX}{\mathbf{X}}

\author{Arnob Ghosh \\
Electrical and Computer Engineering\\
The Ohio State University\\
Columbus, OH, USA \\
\texttt{ghosh.244@osu.edu} 
\And
Xingyu Zhou \\
 Electrical and Computer Engineering\\
Wayne State University\\
Detroit, MI, USA\\
\texttt{xingyu.zhou@wayne.edu} 
\And
Ness Shroff \\
Electrical and Computer Engineering\\
The Ohio State University\\
Columbus, OH, USA \\
\texttt{shroff.11@osu.edu} \\}
 
\title{ Provably Efficient Model-Free Constrained RL with Linear Function Approximation}
\begin{document}
\maketitle
\vspace{-0.2in}
\begin{abstract}
We study the constrained reinforcement learning problem, in which an agent aims to maximize the expected cumulative reward subject to a constraint on the expected total value of a utility function.  In contrast to existing model-based approaches or model-free methods accompanied with a `simulator’, we aim to develop the \emph{first model-free}, \emph{simulator-free} algorithm that achieves a sublinear regret and a sublinear constraint violation even in \emph{large-scale} systems. To this end, we consider the episodic constrained Markov decision processes with linear function approximation, where the transition dynamics and the reward function can be represented as a linear function of some known feature mapping. We show that $\tilde{\mathcal{O}}(\sqrt{d^3H^3T})$ regret and  $\tilde{\mathcal{O}}(\sqrt{d^3H^3T})$ constraint violation bounds can be achieved, where $d$ is the dimension of the feature mapping, $H$ is the length of the episode, and $T$ is the total number of steps. Our bounds are attained without explicitly estimating the unknown transition model or requiring a simulator, and they depend on the state space only through the dimension of the feature mapping. Hence our bounds hold even when the number of states goes to infinity. Our main results are achieved via novel adaptations of the standard LSVI-UCB algorithms. In particular, we first introduce primal-dual optimization into the LSVI-UCB algorithm  to balance between regret and constraint violation. More importantly, we replace the standard greedy selection  with respect to the state-action function in LSVI-UCB with a soft-max policy. 
This turns out to be  key in establishing  uniform concentration for the constrained case via its  approximation-smoothness trade-off. Finally, we also show that one can achieve an even zero constraint violation for large enough $T$ by trading the regret a little bit but still maintaining the same order with respect to $T$.

\end{abstract}

\section{Introduction}


 In many practical applications of online reinforcement learning (RL) (e.g., financial regulations, safety), there exist additional constraints on the learned policy in the sense that it also needs to ensure that the expected total utility (cost, resp.) exceeds a given threshold (is below a threshold, resp.). Constrained RL problem is formulated as a constrained Markov Decision Process (CMDP), in which the celebrated exploration-exploitation trade-off in online RL becomes more challenging due to the additional need to find a balance between reward regret and constraint violations. 

To develop online sample-efficient algorithms for CMDPs, prior works have largely resorted to model-based approaches, where the control policy is constructed based on a learned model~\cite{efroni2020exploration,singh2020learning,brantley2020constrained,zheng2020constrained,kalagarla2020sample,liu2021learning,ding2021provably}. However, due to the explicit estimation and storage of the unknown transition model, model-based approaches often lead to large time and space complexities. These issues are exacerbated in the large state space. Hence, there are several  recent works starting to investigate model-free algorithms for CMDPs, which  directly update the value function or the policy without first estimating the model~\cite{xu2021crpo,ding2020natural,bai2021achieving}. However, all of these works consider an easier setting compared to standard RL in that they assume access to a simulator~\cite{koenig1993complexity} (a.k.a. a generative model~\cite{azar2012sample}), which is a strong oracle that allows the agent to query arbitrary state-action pairs and return the reward and the next state, hence greatly alleviating the intrinsic difficulty of exploration in RL. To the best of our knowledge,~\cite{wei2021provably} were the first to study model-free and simulator-free algorithms for CMDPs.
The above mentioned work on model-free algorithm considers the finite-state tabular setting and the regret scales polynomially with the number of states. Thus, the result {\em would not be useful} for  large-scale RL applications where the number of states could even be infinite.  To address this curse of dimensionality, modern RL has adopted \emph{function approximation} techniques to approximate the (action-)value function \begin{color}{black} of\end{color} a policy, which greatly expands the potential reach of RL, especially via deep neural networks. However, little is known for the performance guarantee of {\em model-free algorithms} in CMDPs beyond tabular settings, even in the case of linear function approximation. 
Motivated by this, we are interested in the following question:
\begin{center}
    \vspace{-1.5mm}
   \emph{Can we achieve provably sample-efficient and model-free exploration for CMDPs beyond tabular settings (without a simulator)?} 
  \vspace{-1.5mm}
\end{center}

\textbf{Contribution.} To answer the above question, we consider the episodic CMDPs
with linear function approximation, where the transition dynamics and the reward function can be represented as a linear function of some known feature mapping. Our main contributions are as follows.
\begin{itemize}[leftmargin=*]
\item We show that with a proper parameter choice, our proposed algorithm achieves  $\tilde{\mathcal{O}}(\sqrt{d^3H^3T})$ regret and  $\tilde{\mathcal{O}}(\sqrt{d^3H^3T})$ constraint violation bounds with a high probability, where $d$ is the dimension of the feature mapping, $H$ is the length of the episode, and $T$ is the total number of steps.  We also show that it is in fact possible to achieve {\em zero} constraint violation by trading the regret a little bit while maintaining the same order with respect to $T$.

\item Our bounds are attained without explicitly estimating the unknown transition model or {\em requiring a simulator}, and they depend on the state space only through the dimension of the feature mapping.  To the best of knowledge, these sub-linear bounds are the first results for model-free, simulator-free online RL algorithms for CMDPs with function approximations. We even improve the bound ($\tilde{\cO}(T^{0.8})$) proposed in the model-free finite state tabular setting by \cite{wei2021provably} (Table~\ref{table:staalgos}).

\item We combine the primal-dual algorithm with the classic model-free, simulator-free LSVI-UCB
algorithm~\cite{jin2020provably} to balance between regret and constraint violations. This naturally leads to the construction of a new composite state-action function (i.e., $Q$-function), which is the sum of the $Q$-function for the reward and the $Q$-function for the utility  weighted by the dual variable. 
Due to this new type of $Q$-function in CMDPs, a key challenge arises when establishing the value-aware uniform concentration, which lies at the heart of the performance analysis of model-free exploration. {\em More specifically, the standard greedy selection with respect to this composite $Q$-function fails in finding non-trivial covering number for the function class of individual value functions (i.e.,  $V$-function) for the reward and the utility respectively.} To address this fundamental issue, we instead adopt a soft-max policy by utilizing its nice property of approximation-smoothness trade-off via its parameter, i.e., temperature coefficient. 

\end{itemize}

 \subsection{Related Work}
 
 Model-based RL algorithms have been proposed for the CMDP \cite{efroni2020exploration,singh2020learning,brantley2020constrained,kalagarla2020sample,liu2021learning, ding2021provably}. Apart from \cite{ding2021provably}, the rest considered tabular set-up. In the tabular model-based set-up, the best known regret and constraint violations achieved are $\tilde{\cO}(\sqrt{|\cS|^2|\cA|T})$ where $|\cS|$ and  $|\cA|$ are the dimensions of the state and action spaces respectively. Hence, such results can not cope up with the large state space. \cite{ding2021provably} considered linear kernel MDP whereas we consider linear MDP. These two are not the same in general.   We  describe the differences with the algorithm proposed in \cite{ding2021provably} in Section~\ref{sec:main_results}. 
 
 Model-free RL algorithms have also been proposed \cite{xu2021crpo,ding2020natural,bai2021achieving} to solve CMDP. However, all of the above require a generator model, which simulates from any state and action.  \cite{wei2021provably} proposed a `triple-Q' algorithm which does not require a 'simulator'. However, it only considered tabular setting.  The regret bound shown in \cite{wei2021provably} is $\tilde{\cO}(T^{0.8})$ which is far from the optimal for the model-based case $\tilde{\cO}(\sqrt{T})$. Please see Table~\ref{table:staalgos} to see our contribution compared to the state-of-the-art approaches.  \cite{amani2021safe} proposed a RL algorithm for the scenario where a constraint needs to be satisfied at each step of an episode. We consider a constraint where the cumulative utility over the length of the episodes must exceed a threshold. Hence, the set of constraints is fundamentally different.  The authors in \cite{amani2021safe} also assumed that a safe-action is known for each state which we do not assume in our setting.  

 
\begin{table*}[!t]
\caption{Regret and Constraint Violations on Episodic MDP for algorithms which do not use simulators. (LFA: Linear Function Approximation, MOD-FREE: Model-Free)}
\label{table:staalgos}
\vspace{-0.1in}
\begin{center}
\begin{small}
\begin{sc}
\begin{threeparttable}
\begin{tabular}{ccccc}
\toprule
Algorithm & LFA? & Mod-free? & Regret & Violations\\
\midrule
  OPDOP \cite{ding2021provably}  & Yes & $\times$ & $\tilde{\cO}(\sqrt{d^2H^5T})$ & $\tilde{\cO}(\sqrt{d^2H^5T})$  \\
  Opt-PrimalDual CMDP \tnote{+} \cite{efroni2020exploration} & No & $\times$ & $\tilde{\cO}(\sqrt{H^5|\cS|^2|\cA|T})$ & $\tilde{\cO}(\sqrt{H^5|\cS|^2|\cA|T})$ \\
  OptDual-CMDP \tnote{+} \cite{efroni2020exploration}  & No & $\times$ & $\tilde{\cO}(\sqrt{H^3|\cS|^2|\cA|T})$ & $\tilde{\cO}(\sqrt{H^3|\cS|^2|\cA|T})$ \\
  OptPress-PrimalDual
  \cite{liu2021learning} & No & $\times$ & $\tilde{\cO}(\sqrt{H^5|\cS|^3|\cA|T})$ & $O(1)$ \\
  Triple-Q \cite{wei2021provably} & No & $\surd$ & $\tilde{\cO}(H^{3.2}T^{0.8}\sqrt{|\cS||\cA|})$ & $0$ \\
  \textbf{Our approach} & Yes & $\surd$ & $\tilde{\cO}(\sqrt{d^3H^3T})$ & $\tilde{\cO}(\sqrt{d^3H^3T})$\tnote{*} \\
\bottomrule
\end{tabular}
\begin{tablenotes}
\item [*] We can reduce the violation to $0$ for large enough $T$ (finite) while maintaining the same order of regret with respect to $T$ (Appendix~\ref{sec:zero_violation}).
\item [+] \small We replace $\rho$ ($\xi$ in our paper) by $\cO(H)$ similar to our paper and $\cN$ by $|\cS|$.
\end{tablenotes}
\end{threeparttable}
\end{sc}
\end{small}
\end{center}
\vskip -0.25in
\end{table*}
\vspace{-0.1in}
\section{Problem Formulation}
We consider an episodic constrained MDP, denoted by $(\cS,\cA,\bP,H,r,g)$ where $\cS$ is the state space, $\cA$ is the action space, $H$ is the fixed length of each episode, $\bP=\{\bP_h\}_{h=1}^{H}$ is a collection of transition probability measures, $r=\{r_h\}_{h=1}^H$ is a collection of reward functions, and $g=\{g_h\}_{h=1}^{H}$ is a collection of utility functions. We assume that $\cS$ is a measurable space with possibly infinite number of elements, $\cA$ is a finite action set. $\bP_h(\cdot|x,a)$ is the transition probability kernel which denotes the probability to reach a state when action $a$ is taken at state $x$. $r_h:\cS\times\cA\rightarrow[0,1]$, and $g_h:\cS\times\cA\rightarrow[0,1]$ and are assumed to be deterministic. However, we can readily extend to settings when $r_h$ and $g_h$ are random.

Each episode $k\in [K]$ starts with the fixed state $x_1$. It can be readily generalized to the setting where $x_1$ is drawn from a distribution. Then at each step $h\in [H]$ in episode $k$, the agent observes state $x_h^k\in \cS$, picks an action $a_h^k\in \cA$, receives a reward $r_h(x_h^k,a_h^k)$, and a  utility $g_h(x_h^k,a_h^k)$. The MDP evolves to $x_{h+1}^k$ that is drawn from $\bP_h(\cdot|x_h^k,a_h^k)$. The episode terminates at step $H+1$. Without loss of generality, we assume that $r_{H+1}=g_{H+1}=0$. In this paper, we consider the {\em challenging} scenario where the agent only observes the bandit information $r_h(x_h^k,a_h^k)$ and $g_h(x_h^k,a_h^k)$ at the visited state-action pair $(x_h^k,a_h^k)$.
The policy-space of an agent is  $\Delta (\cA|\cS,H)$; $\{\{\pi_h(\cdot|\cdot)\}_{h=1}^H: \pi_h(\cdot|x)\in \Delta(\cA), \forall x\in \cS,$ and $h\in [H]$\}. Here $\Delta(\cA)$ is the probability simplex over the action space. For any $x_h^k\in \cS$ , $k\in [K]$, and $h\in [H]$, $\pi_{h,k}(a_h^k|x_h^k)$ denotes the probability that the action $a_h^k\in \cA$ is taken at episode $k$ when the state is $x_h^k$. 

Let $V^{\pi}_{r,h}(x)$ denote the expected value of the total reward function starting from step $h$ and state $x$ when the agent selects action using the policy $\pi=\{\pi_{h}\}_{h=1}^{H}$
\begin{align}
   V^{\pi}_{r,h}(x)= \bE_{\pi}\left[\sum_{i=h}^{H}r_i(x_i,a_i)|x_h=x\right],
\end{align}
where $\bE$ is taken with respect to the policy $\pi$ and the transition probability kernel $\bP$. Let $Q^{\pi}_{r,h}(x,a)$ denote the expected value of the total reward starting from step $h$ and the state-action pair $(x,a)$ and follows the policy $\pi$ as
\begin{align}
    Q^{\pi}_{r,h}(x,a)=\bE_{\pi}\left[\sum_{i=h}^{H}r_i(x_i,a_i)|x_h=x,a_h=a\right].
\end{align}
Similarly, we define the value function for the utility $V^{\pi}_{g,h}(x)$, and the action-value function for the utility $Q^{\pi}_{g,h}(x,a)$.  We denote $V^{\pi}_{j,h}(x)$, and $Q^{\pi}_{j,h}(x,a)$ for $j=r,g$. 
\begin{definition}
For brevity, we denote $\bP_hV^{\pi}_{j,h+1}(x,a)=\bE_{x^{\prime}\sim\bP_h(\cdot|x,a)}V^{\pi}_{j,h+1}(x^{\prime})$ for $j=r,g$.
\end{definition}
Using this notation, the Bellman's equation associated with the policy $\pi$ becomes
\begin{align}\label{eq:bellman}
    Q^{\pi}_{j,h}(x,a)=(r_h+\bP_hV^{\pi}_{j,h+1})(x,a)
\end{align}
Note that $V^{\pi}_{j,h}(x)=\langle\pi_h(\cdot|x),Q^{\pi}_{j,h}(x,\cdot)\rangle_{\cA}$, where $\langle\pi_h(\cdot|x),Q^{\pi}_{j,h}(x,\cdot)\rangle_{\cA}=\sum_{a\in \cA}\pi_h(a|x)Q^{\pi}_{j,h}(x,a)$.

The objective of the learning agent is to find an optimal solution of the following problem
\begin{align}\label{eq:cmdp}
& \text{maximize }_{\pi\in \Delta(\mathcal{A}|\mathcal{S},\mathcal{H})} V^{\pi}_{r,1}(x_1),\quad
& \text{subject to } V^{\pi}_{g,1}(x_1)\geq b.
\end{align}
Note that even though we have only once constraint, it can be readily generalized to the scenario with multiple constraints. In order to avoid trivial solutions, we consider $b\in (0,H]$. We denote the optimal policy as $\pi^{*}$ which solves the above optimization problem. Since $\pi^{*}$ is obtained by having complete information, it is also denoted as {\em the best policy in the hindsight}.

Without any constraint information a priori, an agent can not know the policies that satisfy the constraint. Instead, we allow the policy to violate the constraint and minimize the regret while minimizing the total constraint violations over the $K$ episodes. Such an approach is also considered in the existing literature \cite{efroni2020exploration,ding2021provably,ding2020natural}. We now define the performance metric which we seek to minimize.

\textbf{Performance Metric.} Let the policy employed by the agent at episode $k$ be $\pi_k=[\pi_{1,k},\ldots,\pi_{h,k},\ldots,\pi_{H,k}]^T$. The performance metric we are considering is the following 
\begin{align}
\text{Regret}(K)=\sum_{k=1}^{K}V^{\pi^*}_{r,1}(x_1)-V^{\pi_k}_{r,1}(x_1)\nonumber\\
\text{Violation}(K)=\left[\sum_{k=1}^{K}(b-V^{\pi_k}_{g,1}(x_1))\right]_{+},
\end{align}
where $[z]_{+}=\max\{z,0\}$. The regret is defined as the difference between the total reward value by following the optimal policy $\pi^{*}$, and the total reward value obtained by following agent's policy $\pi_k$ at episode $k$ over $K$ episodes. The constraint violation is defined as the difference between the threshold value $Kb$ and the total utility function attained by following the policies over all the episodes $K$.

\textbf{Linear Function Approximation.} To handle a possible large number of states, we consider the following linear MDPs. 
\begin{assum}\label{assum:linearmdp}
The CMDP is a linear MDP with  feature map $\phi: \mathcal{S}\times\mathcal{A}\rightarrow\R^{d}$, if for any h, there exists $d$ {\em unknown} signed measures $\mu_h=\{\mu^1_h,\ldots,\mu^d_h\}$ over $\mathcal{S}$  such that for any $(x,a,x^{\prime})\in \mathcal{S}\times \mathcal{A}\times \mathcal{S}$, 
\begin{align}
\bP_h(x^{\prime}|x,a)=\langle\phi(x,a),\mu_h(x^{\prime})\rangle
\end{align}
and there exists vectors $\theta_{r,h}$, $\theta_{g,h}\in \R^d$ such that for any $(x,a)\in \mathcal{S}\times\mathcal{A}$,
\begin{align*}
r_h(x,a)=\langle\phi(x,a),\theta_{r,h}\rangle\quad g_h(x,a)=\langle\phi(x,a),\theta_{g,h}\rangle
\end{align*}
\end{assum}
\vspace{-0.1in}
Assumption~\ref{assum:linearmdp} adapts the definition of linear MDP \cite{jin2020provably, yang2019sample} to the constrained case. By the above definition, the transition model, the reward, and the utility functions are linear in terms of feature map $\phi$. We remark that despite being linear, $\bP_h(\cdot|x,a)$ can still have infinite degrees of freedom since $\mu_h(\cdot)$ is unknown. Note that tabular MDP is part of linear MDP \cite{jin2020provably}.

Note that \cite{ding2021provably,zhou2021provably} studied another related concept known as linear kernel MDP. In the linear kernel MDP, the transition probability is given by $\bP_h(x^{\prime}|x,a)=\langle\psi(x^{\prime},x,a),\theta_h\rangle$. In general, linear MDP and linear kernel MDPs are two different classes of MDP \cite{zhou2021provably}. 


Similar to Proposition 1 in \cite{jin2020provably}, we can show that for a linear MDP and for any policy $\pi$ there exists $\{w_{j,h}^{\pi}\}_{h=1}^H$ such that $Q^{\pi}_{j,h}(x,a)=\langle w_{j,h}^{\pi},\phi(x,a)\rangle$ for any $(x,a,h)\in \cS\times\cA\times[H]$.  We, thus, focus on linear action-value function. 

\textbf{Dual problem and Slater's Condition.}
We first introduce few notations which we will use throughout this paper. 
\begin{definition}\label{defn:lagrange_value}
$V^{\pi,Y}_{h}(\cdot)=V^{\pi}_{h,r}(\cdot)+Y V^{\pi}_{h,g}(\cdot)$, and $Q^{\pi,Y}_h(x,a)=Q^{\pi}_{r,h}(x,a)+YQ^{\pi}_{g,h}(x,a)$, where $Y$ is the dual variable. 
\end{definition}

Thus, $V^{\pi,Y}_h(\cdot)$ and $Q^{\pi,Y}_h(\cdot,\cdot)$ are respectively the composite value function and $Q$-functions respectively. 
We can cast the problem (\ref{eq:cmdp}) as a saddle point problem $\max_{\pi}\min_{Y}\cL(\pi,Y)$ where $\cL(\pi,Y)=V_{r,1}^{\pi}(x_1)+Y(V_{g,1}^{\pi}(x_1)-b)=V_1^{\pi,Y}-Yb$, where $\pi$ is the primal policy and $Y$ is the dual variable. However, the lagrangian is non-concave in $\pi$ \cite{agarwal2021theory} even though it is convex in $Y$. Nevertheless, the strong duality holds \cite{paternain2019safe}. Hence, there exists optimal dual variable $Y^{*}$, such that $\max_{\pi}\cL(\pi,Y^*)$ will correspond to the optimal reward value function. 

We assume the following slater's condition in this paper.
\begin{assum}[Slater's Condition]\label{assum:slater}
There exists $\gamma>0$, and $\bar{\pi}\in \Delta(\mathcal{A}|\mathcal{D},\mathcal{H})$, such that $V^{\bar{\pi}}_{g,1}(x_1)\geq b+\gamma$,
\end{assum}

\begin{lem}[Boundedness of $Y^{*}$]\label{lem:boundY}
The optimal dual-variable $Y^{*}\leq \dfrac{V^{\pi^*}_{r,1}(x_1)-V^{\bar{\pi}}_{r,1}(x_1)}{\gamma}\leq \dfrac{H}{\gamma}$. 
\end{lem}


The slater's condition is mild in practice and commonly adopted in previous works \cite{ding2021provably,efroni2020exploration,qiu2020upper}. We use the properties of the slater's condition to bound the performance of our proposed algorithm. 



\begin{definition}\label{defn:xi}
We set $\xi=2H/\gamma$.
\end{definition}


\section{Our Approach}
\begin{algorithm}[h]
%
	\caption{Model Free Primal-Dual Algorithm for Linear Function Approximation} 
	\label{algo:model_free}
	\begin{algorithmic}[1]
	%
			\STATE \textbf{Initialization: } $Y_1=0$, $w_{j,h}=0$, $\xi=2H/\gamma$,$\alpha=\dfrac{\log(|\cA|)K}{2(1+\xi+H)}$, $\eta=\xi/\sqrt{KH^2}$, $\beta=C_1dH\sqrt{\log(4\log|\cA| dT/p)}$
		
			\FOR {episodes $k=1,\ldots, K$}
		\STATE Receive the initial state $x_1^k$.
		\FOR {step $h=H,H-1,\ldots, 1$}
		\STATE $\Lambda^k_{h}\leftarrow \sum_{\tau=1}^{k-1}\phi(x_h^{\tau},a_h^{\tau})\phi(x_h^{\tau},a_h^{\tau})^T+\lambda \bI$
\STATE $w^k_{r,h}\leftarrow (\Lambda^k_h)^{-1}[\sum_{\tau=1}^{k-1}\phi(x_h^{\tau},a_h^{\tau})[r_h(x_h^{\tau},a_h^{\tau})+V^k_{r,h+1}(x_{h+1}^{\tau})]]$
\STATE$w^k_{g,h}\leftarrow (\Lambda^k_{h})^{-1}[\sum_{\tau=1}^{k-1}\phi(x_h^{\tau},a_h^{\tau})[g_h(x_h^{\tau},a_h^{\tau})+V^k_{g,h+1}(x^{\tau}_{h+1})]]$
\STATE $Q^k_{r,h}(\cdot,\cdot)\leftarrow \min\{\langle w_{r,h}^k,\phi(\cdot,\cdot)\rangle+\beta(\phi(\cdot,\cdot)^T(\Lambda^k_{h})^{-1}\phi(\cdot,\cdot))^{1/2},H\}$
\STATE $Q^k_{g,h}(\cdot,\cdot)\leftarrow \min\{\langle w_{g,h}^k,\phi(\cdot,\cdot)\rangle+\beta(\phi(\cdot,\cdot)^T(\Lambda^k_{h})^{-1}\phi(\cdot,\cdot))^{1/2},H\}$
\STATE $\pi_{h,k}(a|\cdot)=\dfrac{\exp(\alpha(Q^k_{r,h}(\cdot,a)+Y_kQ^k_{g,h}(\cdot,a)))}{\sum_{a}\exp(\alpha(Q^k_{r,h}(\cdot,a)+Y_kQ^k_{g,h}(\cdot,a)))}$
\STATE $V^k_{r,h}(\cdot)=\sum_{a}\pi_{h,k}(a|\cdot)Q^k_{r,h}(\cdot,a)$
\STATE $V^k_{g,h}(\cdot)=\sum_{a}\pi_{h,k}(a|\cdot)Q^k_{g,h}(\cdot,a)$

		\ENDFOR
		
		
		\FOR {step $h=1,\ldots,H$}
		\STATE Compute $Q_{r,h}^k(x_h^k,a)$, $Q_{g,h}^k(x_h^k,a)$, $\pi(a|x_{h}^k)$ for all $a$.
\STATE Take action $a_h^k\sim \pi_{h,k}(\cdot|x_h^k)$ and observe $x_{h+1}^k$. 
\ENDFOR
\STATE $Y_{k+1}=\max\{\min\{Y_k+\eta(b-V^k_{g,1}(x_1)),\xi\},0\}$
\ENDFOR
	
	\end{algorithmic}
\end{algorithm}
\vspace{-0.1in}

 We now describe our proposed algorithm in Algorithm~\ref{algo:model_free}. This algorithm is based on the primal-dual adaptation of the LSVI-UCB \cite{jin2020provably}. For a given dual variable, the primal policy is updated, and then the dual value is updated based on the estimated utility value function. At each episode, the algorithm consists of three parts. The first part (Steps 4-12) consists of updating the parameters $w^k_{r,h},w^k_{g,h}$ and $\Lambda^k_h$ which are used to update the $Q^k_{j,h}$  and $V^{k}_{j,h}$ at episode $k$.  $\Lambda_h^k$ is the Gram-matrix for the regularized least square problem (see Eqn. (\ref{eq:ls}), later). Note that the Steps 8-12 are not evaluated for each state, rather, they are evaluated only for the encountered states till episode $k-1$. Hence, we do not need to iterate over potentially infinite number of states. For the first episode, since $k-1=0$ and $\tau=1$, we have $w^k_{j,h}=0$, $\forall j$ and $\Lambda^k_h=\lambda\bI$. We note that $Q^{k}_{j,H+1}(\cdot,\cdot)=0$ for $j=r,g$. 
 
 The value functions are updated (Steps 11-12) based on $Q$ function and the policy. The policy is based (Step 10) on a soft-max policy unlike the greedy one in the unconstrained case~\cite{jin2020provably}. Soft-max policy $\textsc{Soft-Max}_{\alpha}(\vX)=\{\textsc{Soft-Max}^i_{\alpha}(\vX)\}_{i=1}^{|\cA|}$ for any vector $\vX\in \R^{|\cA|}$ is a $|\cA|$-dimensional vector with parameter $\alpha$ where the $i$-th component
\vspace{-0.05in}
\begin{align}\label{eq:soft-max}
    \textsc{Soft-Max}^i_{\alpha}(\vX)=\dfrac{\exp(\alpha X_i)}{\sum_{n=1}^{|\cA|}\exp(\alpha X_n)}
\end{align}
 At step $h$, $\pi_{h,k}(a|x)$ is computed based on the soft-max policy on the composite $Q$-function vector $\{Q^k_{r,h}(x,a)+Y_kQ^k_{g,h}(x,a)\}_{a\in\cA}$ where $Y_k$ is the lagrangian multiplier. When $\alpha=\infty$, this becomes equal to the greedy policy.  The second part (Steps 13-15) is the execution of the soft-max policy based on the composite $Q$-value for the encountered state $x_h^k$.


\textbf{$Q$ function and Value function Estimation.} 
We need to estimate the value-function and $Q$-function with respect to the policy $\pi_k$. However, there are challenges. We do not know $\bP_h$ in Bellman's equation (\ref{eq:bellman}), rather $\bP_hV^{\pi_k}_{j,h+1}$ should be replaced by the empirical samples. Further, in the large state space, we can not iterate over all $(x,a)$. Rather, we parameterize  $Q_{j,h}^{\pi^*}(\cdot,\cdot)$ by a linear form $\langle w_{j,h}^k(\cdot,\cdot),\phi(\cdot,\cdot)\rangle$. The intuition is to obtain $w_{j,h}^k$ from the Bellman's equation using the regularized least-square regression. We obtain $w_{j,h}^k$ for $j=r,g$ according to the following equation
\vspace{-0.1in}
\begin{align}\label{eq:ls}
    w_{j,h}^k\leftarrow \arg\min_{w\in \R^{d}}& \sum_{\tau=1}^{k-1}[j_h(x_h^{\tau},a_h^{\tau})+V_{j,h+1}^k(x_{h+1}^{\tau}) -w^T\phi(x_h^{\tau},a_h^{\tau})]^2 +\lambda ||w||_2^2
\end{align}
Then, an additional bonus term $\beta(\phi(\cdot,\cdot)^T(\Lambda^k_h)^{-1}\phi(\cdot,\cdot))^{1/2}$ is added as in \cite{jin2020provably}. $\beta$ is constant which we will characterize in the next section.  Such an additional term is used for upper confidence bound in LSVI-UCB \cite{jin2020provably}. The same additional term is used for both $Q^k_{r,h}$ and $Q^k_{g,h}$. Note the difference with the LSVI-UCB, here, we need to estimate the value function corresponding to the soft-max policy $\pi_k$ where in LSVI-UCB, a greedy policy corresponding to the $Q$-function is used. 

\textbf{Policy.} We update a soft-max policy which selects actions according to the estimated `composite'  $Q$-function at the $k$-th episode. The reason behind using the soft-max policy instead of a greedy policy will be apparent in the next section when we state the main results and the proof ideas. Note from the strong duality, for optimal dual variable $Y^{*}$, optimal primal policy $\pi^{*}$ maximizes the composite value function $V^{\pi^*,Y^*}$ (Definition~\ref{defn:lagrange_value}). Thus, the optimal policy should be a greedy one based on this optimal dual value $Y^*$. However, the greedy policy is not Lipschitz, hence, it does not provide uniform concentration bound for each individual value function, an essential step in the regret bound (Section~\ref{sec:outline}). Hence, compared to the unconstrained scenario, we need more exploration in the policy space  where the apparent reason is that we do not know the optimal dual variable beforehand. Since we use the soft-max policy, there is a gap compared to the optimal value even when the lagrangian multiplier $Y_k$ becomes equal to $Y^{*}$.  However, if $\alpha$ in the soft-max policy also scales with $K$, then we can bound the gap from the optimal value function (Section~\ref{sec:outline}). 



\textbf{Dual Update.} To infer the constraint violation, we estimate $V_{g,1}^{k}$ for $V_{g,1}^{\pi_k}$. We update the lagrangian multiplier $Y_k$ by moving towards minimizing the lagrangian $\cL(\pi,Y)$ over $Y\geq 0$ in line 16, where $\eta>0$ is a step-size and $\xi$ is the upper bound on the dual variable such that optimal dual variable $Y^{*}$ is contained within $[0,\xi]$.  The dual update is similar to the step described in \cite{ding2021provably,efroni2020exploration}.

The dual update works as a trade-off between the reward maximization and the constraint violation reduction. If $b-V_{g,1}^k\geq 0$, that means  with a high probability, the constraint will be violated for the policy $\pi_k$. Hence, the dual value is increased in order to focus on minimizing the constraint violation. Otherwise, the agent tries to maximize the reward value function.

\textbf{Space and Time Complexities.}   We  remark that Algorithm~\ref{algo:model_free} only needs to store $Y_k$, $w_{r,h}^k,w_{g,h}^k,r_h(x_h^k,a_h^k), g_h(x_h^k,a_h^k),\Lambda_h^k$, and  $\{\phi(x_h^k,a)\}_{a\in\cA}$ for all $(h,k)\in [H]\times[K]$, hence, it takes $\cO(d^2H+d\cA T)$ space. When we compute  $(\Lambda_h^k)^{-1}$ using Sherman-Morrison formula, the computation of $V_{j,h+1}^k$ is dominated by computing $Q_{j,h+1}^k$ and the policy $\pi_k$. Hence, it takes $\cO(d^2\cA T)$ time. Note that since our approach is model-free and we do not need to evaluate integrals as in \cite{ding2021provably} in order to estimate $\bP_hV_{h+1}^k$.

Note that both $\eta$ and $\alpha$ use the knowledge of $K$. In case, $K$ is unknown, one can use the ``doubling trick" \cite{besson2018doubling} which will only scale the regret and constraint violation by a constant factor. 

\section{Analysis}\label{sec:analysis}
We now state the main result. We prove that Algorithm~\ref{algo:model_free} achieves regret and constraint violation which are sublinear in $T=KH$ where $T$ is the total number of steps. 
\subsection{Main Results}\label{sec:main_results}
\begin{theorem}\label{thm:episodic}
Fix $p>0$. If we set $\lambda=1$, $\beta=C_1dH\sqrt{\iota}$ in Algorithm~\ref{algo:model_free} where $\iota=\log(\log(|\cA|)4dT/p)$ for some absolute constant $C_1$. With probability $(1-p)$, 
\begin{align}
& \mathrm{Regret}(K)\leq C\sqrt{d^3H^3T\iota^2}+\xi\sqrt{HT}\nonumber\\
& \mathrm{Violation}(K)\leq \dfrac{C^{\prime}2(1+\xi)}{\xi}\sqrt{d^3H^3T\iota^2}\nonumber
\end{align}
for some absolute constants $C$, and $C^{\prime}$.
\end{theorem}


We remark the difference with the existing results. Since $\xi=2H/\gamma$ (by Definition~\ref{defn:xi}), our result indicates that our approach obtains $\mathcal{\tilde{O}}(\sqrt{d^3H^3T})$ regret and the same order of constraint violation where $\tilde{\cO}$ absorbs logarithmic factor on $T$.  The regret and constraint violation are sub-linear in $T$, and similar dependence is observed in \cite{efroni2020exploration, ding2021provably}. Also note that compared to the unconstrained case \cite{jin2020provably}, there is an additional $\log(|\cA|))$ factor in the value of $\iota$ which arises because we use soft-max policy instead of the greedy policy which adds to the covering number. The regret and constraint violation do not depend on the dimension of the state space, rather, it depends on the dimension of the feature space. {\em To the best of our knowledge this is the first result which shows both $\tilde{\cO}(\sqrt{T})$ regret and constraint violation in the model-free set up (tabular or linear) without requiring a simulator. }

\textbf{Comparison with \cite{ding2021provably}}: Compared to \cite{ding2021provably} which also considers linear function approximation (however, it considers linear kernel MDP rather linear MDP) we improve the result in \cite{ding2021provably} by a factor of $H$.
Second, compared to \cite{ding2021provably}, which is a model-based policy-based algorithm, ours is a model-free value-based algorithm. Due to this, the above uniform concentration challenge does not exist in \cite{ding2021provably}. Moreover, our model-free algorithm also enjoys an easy implementation and improved computation efficiency since it does not estimate the next step expected value function as in \cite{ding2021provably} which requires an  integration oracle to compute a $d$-dimensional integration at every step. \cite{ding2021provably} also needs to store the previous policies and estimated value functions, hence, it needs $O(T)$ additional space complexity.  
We have an additional $\sqrt{d}$ factor in front of the regret and constraint violation. Similar difference  in regret is also observed between the model-based linear kernel unconstrained MDP \cite{ayoub2020model} and model-free linear unconstrained MDP \cite{jin2020provably} \emph{even in the unconstrained case}. 

Similar to the discussion in Section 3.1 on \cite{jin2018q}, our result directly translates to a sample complexity guarantee (or, PAC guarantee).  For example, we  can learn a policy $\pi$ such that $V^{\pi^*}_{r,1}(x_1)-V^{\pi}_{r,1}(x_1)\leq \epsilon$, and $b-V^{\pi}_{g,1}(x_1)\leq \epsilon$ after $\tilde{\cO}(d^3H^4/\epsilon^2)$ number of samples. Here, the policy $\pi$ is obtained after running Algorithm~\ref{algo:model_free} for $\tilde{\cO}(d^3H^3/\epsilon^2)$ number of episodes, and then selecting policy $\pi_k$ with probability $1/K$ for any $k\in [K]$. 

Recently, \cite{miryoosefi2022simple} proposed an algorithm with provable sample complexity guarantee for linear CMDP. However, the regret and violation guarantees are different from the sample complexity guarantees as the former ones are {\em any time} guarantee. The proposed algorithms are  different since the goal is different. In particular, the uniform concentration bound challenge does not appear there.  Note that using the explore-then-commit algorithm \cite{jin2018q}, one can achieve $\tilde{\cO}(T^{2/3})$ regret for large $T$ (from $\tilde{\cO}(1/\epsilon^2)$ sample complexity bound achieved in \cite{miryoosefi2022simple}) which is worse than ours. Additionally, we achieve zero violation (Remark~\ref{rem:zero_violation}) while maintaining the same order of regret with respect to $T$.

\subsection{Outline of the Proof}\label{sec:outline}
In this section, we provide an outline of our proof, which is mainly divided into three steps. We first establish a decomposition of the sum of reward regret and constraint violation. Then, we will bound two key terms that are related to optimism and prediction error, respectively. Finally, using standard optimization tools, we can achieve the main results. We highlight that the key challenges lie in the second step where a balance between the optimistic term and prediction error term is handled via the introduced soft-max policy. 

\textbf{Step 1: Bounding the sum of Regret and violation scaled by dual variable} Similar to~\cite{efroni2020exploration}, we first establish the following decomposition, which upper bounds the sum of regret and  violation. This will serve as the basis when applying optimization tools in Step 3.

\begin{lem}[Decomposition] 
\label{lem:dual_variable}
For any $Y \in [0,\xi]$, we have 
\begin{align*}
\label{eq:regret}
    &\sum_{k=1}^K (V_{r,1}^{\pi^*}(x_1) - V_{r,1}^{\pi_k}(x_1)) + Y \sum_{k=1}^K (b-V_{g,1}^{\pi_k}(x_1)) 
     \le \frac{1}{2\eta} Y^2 + \frac{\eta}{2}H^2K + \nonumber\\& \underbrace{\sum_{k=1}^K \left(V_{r,1}^{\pi^*}(x_1) + Y_kV_{g,1}^{\pi^*}(x_1) \right) - \left(V_{r,1}^k(x_1) + Y_k V_{g,1}^k(x_1)\right)}_{\mathcal{T}_1}
    +\nonumber\\ &  \underbrace{\sum_{k=1}^K \left(V_{r,1}^k(x_1) - V_{r,1}^{\pi_k}(x_1)\right) + Y \sum_{k=1}^K \left(V_{g,1}^k(x_1) - V_{g,1}^{\pi_k}(x_1)\right)}_{\mathcal{T}_2}
\end{align*}
\end{lem}
Note that $\cT_1$ is similar to the term related to optimism in the unconstrained case with the difference being that we now have two value functions weighted by the dual variable $Y_k$. Similarly, $\cT_2$ is similar to prediction error term with the additional weight by $Y$. Since the first term in the above inequality can be easily bounded with a proper choice of $\eta$, we are only left to bound $\cT_1$ and $\cT_2$, respectively. 

\textbf{Step 2: Bounding $\cT_1$ and $\cT_2$} To bound $\cT_2$ and $\cT_1$, we need to bound the difference between the \emph{individual} estimated value function $V_{j,h}^k$ and the \emph{individual} value function $V_{j,h}^{\pi}$ corresponding to a given policy $\pi$ at episode $k$. As in the unconstrained case, the key step is to control the  fluctuations in least-squares value iteration. In particular, we need to show that for all $(k,h)\in [K]\times [H]$ with high  probability
\begin{equation*}\label{eq:martin_p}
\norm{\sum_{\tau=1}^{k-1}\phi(x_h^{\tau},a_h^{\tau}) \left[V^k_{j,h+1}(x_{h+1}^{\tau})-\bP_hV^k_{j,h+1}(x_{h}^{\tau},a_h^{\tau})\right]}_{(\Lambda^k_h)^{-1}}
\end{equation*}
is upper bounded by lower order term (e.g., $\cO(d\sqrt{\log K})$. To this end, value-aware uniform concentration is required to handle the dependence between $V_{j,h+1}^k$ and samples $\{x_{h+1}^{\tau}\}_{\tau=1}^{k-1}$, which renders the standard self-normalized inequality infeasible in the model-free setting. The general idea here  is to fix a function class $\cV_{j,h}$ in advance and then show that each possible value function in our algorithm $V_{j,h}^k$ is within this class which has polynomial log-covering number. 
\emph{In the following, we fix an  $h \in [H]$ and drop the subscript $h$ for notation simplicity.}

\textbf{Uniform Concentration Bound for class of value function:}
We first note that this uniform concentration bound is the main motivation for us to choose a soft-max policy as we will see that the standard greedy policy would fail in this case. That is, in order to guarantee that for each possible $V_j^k$, there is an $\epsilon$-close function in $\mathcal{V}_j$, it would basically lead to a very large covering number. To address this, we introduce soft-max policy and define the following corresponding function classes. 
We first define the following class for $Q$-function for $j = r,g$.
 $\cQ_j=\{Q_j|Q_j(\cdot,\cdot)=
\min\{\langle w_{j},\phi(\cdot,\cdot)\rangle
 +\beta\sqrt{\phi(\cdot,\cdot)^T(\Lambda_h)^{-1}\phi(\cdot,\cdot)},H\}.$
Then, we define the following value function class $\cV_j$.
   $ \cV_j = \{V_j| V_j(\cdot) = \sum_a \pi(a|\cdot) Q_j(\cdot,a); Q_j \in \cQ_j, \pi \in \Pi\},$
where $\Pi$ is given by the following class 
 $  \Pi =\{\pi|\pi(a|\cdot) = \textsc{Soft-Max}^a_{\alpha}((Q_{r}(\cdot,\cdot)+Y Q_{g}(\cdot,\cdot));
  \forall a\in \cA, Q_r \in \cQ_r, Q_g \in \cQ_g, Y \in [0,\xi]\},$
where $\textsc{Soft-Max}$ is defined in (\ref{eq:soft-max}).


\textbf{ Why soft-max?}
At this moment, we can explain why the introduction of soft-max in our algorithm is critical. Suppose we follow the standard greedy selection, which corresponds to $\alpha = \infty$ in above. The key issue in this approach is that one needs a  large  $\epsilon$-covering for $\cV_j$ so that each possible $V_j^k$ can be well-approximated (i.e., $\epsilon$-close) by function in $\cV_j$. This is in sharp contrast to the unconstrained case where $\cV_j$ has a polynomial  log-covering number. To see this difference, in the unconstrained case, we only have $V_r^k$ that is greedy with respect to $Q_r^k$. By the fact that $\max_a$ is a contraction map, an $\epsilon$-covering of the $Q_r^k$ implies an $\epsilon$-covering of $V_r^k$ and meanwhile the covering number of $\cQ_r$ is reasonably small  (Lemma~\ref{lm:qcovering}) by standard arguments. This no longer holds in the constrained case due to the use of a composite $Q$-function. In particular, note that if the policy is greedy w.r.t. the composite $Q$-function, then an $\epsilon$-covering of $Q_j^k$ {\em fails} to be an $\epsilon$-covering of $V_j^k$ in general since the greedy policy is not smooth in that a slight change of the composite $Q$-function could lead to a substantial change of the output action. This leads to a large  distance for \emph{individual} value functions due to the different action choices, even though the $Q$-function is close (Please see Appendix~\ref{sec:nogreedy} for an example). Hence, one can not approximate individual value function within $\epsilon$-bound using greedy policy based on composite $Q$-function.
This fact motivates us to turn to $\textsc{Soft-Max}_{\alpha}$, which is Lipschitz continuous with a Lipschitz constant at most $2\alpha$. Thus, our main idea is as follows. Given $Q_r^k$, $Q_g^k$ and $Y_k$, we can first find fixed $\tilde{Q}_r \in \cQ_r$ , $\tilde{Q}_g \in \cQ_g$ and $\tilde{Y} \in [0,\xi]$ such that $\norm{Q_r^k -\tilde{Q}_r}_{\infty} \le \epsilon_1$, $\norm{Q_g^k -\tilde{Q}_g}_{\infty} \le \epsilon_2$,  $\norm{Y^k -\tilde{Y}}_{\infty} \le \epsilon_3$ and $\norm{(Q_r^k + Y_k Q_g^k) - (\tilde{Q}_r + \tilde{Y} \tilde{Q}_g)}_{\infty} \le \epsilon$ with a reasonably small covering number. Then, thanks to the smoothness of soft-max function, we have $\norm{\pi_k - \tilde{\pi}}_1 \le 2\alpha \epsilon$ (Lemma~\ref{lem:pi}). Combining this with the closeness of individual $Q$-function yields the closeness of individual $V$-function (Lemma~\ref{cor:vcovering}). Hence, it ensures that the class $\cV_j$ in our set-up has  log-covering number of $\cO(d\log(K))$. 

\textbf{Choosing hyper-parameter $\alpha$ to achieve bound:} A larger value of $\alpha$ means that we need a smaller $\epsilon$, hence a larger covering number. 
Then, one may wonder if  we can choose an arbitrarily small value for $\alpha$. However, the term $\cT_1$ will be enlarged if we choose too small $\alpha$. Note that in the unconstrained case, $\cT_1$ is upper bounded by zero due to optimism under greedy policy. Now, since we are using soft-max, we need to bound the approximation error between the soft-max and greedy one. As expected, in this case, a larger $\alpha$ leads to a smaller approximation error. 

From the discussions above, we can see that the approximation-smoothness trade-off of the soft-max function is well captured by our $\cT_1$ and $\cT_2$, respectively. Therefore, we need to carefully choose the value of $\alpha$ to balance these two. In particular, with $\alpha = \frac{\log(|\cA|)K}{2H}$, we have the following bounds on $\cT_1$ and $\cT_2$, respectively. 

\begin{lem}\label{lem:ucb}
With probability $1-p/2$, we have $\mathcal{T}_1\leq K\dfrac{H\log(|\cA|)}{\alpha}$. Hence, for $\alpha=\dfrac{\log(|\cA|)K}{2(1+\xi+H)}$, we have $\mathcal{T}_1\leq 2H(1+\xi+H)$ with probability $1-p/2$.
\end{lem}

 \begin{lem}\label{lem:conc}
With probability at least $1-p/2$,
$\mathcal{T}_2\leq \mathcal{O}((Y+1)\sqrt{d^3H^3T\iota^2})$, where $\iota=\log[\log(|\cA|)4dT/p]$
\end{lem}
\begin{remark}
The additional $\log |A|$ factor in $\iota$ arises as a trade-off for selecting soft-max policy as it is evident in Lemma~\ref{lem:ucb}. When we compensate by making $\alpha$ scaled with $\log|A|$ in Lemma~\ref{lem:ucb}, it increases the covering-number by $\log|A|$ as well in the $\iota$ term.  
\end{remark}

\textbf{Step 3: Final Result by combining all the pieces:} 
By replacing $Y=0$, $\eta=\xi/(\sqrt{KH^2})$, and combining Lemma~\ref{lem:dual_variable},\ref{lem:ucb}, and \ref{lem:conc}, we obtain the regret bound.  We also obtain the constraint violation using the idea from \cite{efroni2020exploration}. 

\begin{remark}\label{rem:zero_violation}

We can reduce the violation to zero while maintaining the same order on regret with respect to $T$ (Appendix~\ref{sec:zero_violation}). We consider a tighter optimization problem where we add $\zeta$ in the constraint of (\ref{eq:cmdp}). In Appendix~\ref{sec:zero_violation}, we bound the difference  between the optimal value function for the tighter and the original problem as a function of $\zeta$. Since the tighter problem is also CMDP, we attain the regret and violation bound as in Theorem~\ref{thm:episodic} with $b+\zeta$ in place of $b$.  Hence, by choosing $\zeta$, we can show that it is possible to achieve $\tilde{\cO}(\sqrt{T})$ regret and zero violation for large enough $K$ (Theorem~\ref{thm:zero_violation} in Appendix~\ref{sec:zero_violation}) albeit with an extra $H$ factor in front of regret bound. 
%



\end{remark}
\section{Experiments}
We evaluate Algorithm~\ref{algo:model_free} on a simulated model for job scheduling to validate our theoretical results. We consider that the number of jobs belongs to the discrete state $\{0,1,\ldots,9\}$ where $0$ means that there is no job. Total time horizon ($H$) is divided in $10$ steps. After $H$ steps, a new episode begins. At the start of the each episode, the state of the job is assumed to be $9$, i.e., the job stack is full.  The agent  needs to decide whether to send job $(a=1)$ or not ($a=0$) to a machine. We assume that if $a=1$, the agent sends $2$ jobs to a machine and incurs a cost as the machine spends some resources to process the job. We assume that the rewards are step dependent. In particular, we assume that at time steps from $3$ to $6$, the reward is $1-0.9a$, In other time steps, the reward is $1-0.2a$. This mimics the setup where at certain time, it might be more costly to process a job (for example, electricity cost might be higher, or the machine needs to abandon an important job).

We assume that even if the scheduler schedules a job, the machine might not be able to complete the $2$ jobs. In particular, 
\begin{align}
    x_{h+1}=\begin{cases} \max\{x_h-2a,0\} \quad \text{w.p. } 0.8\nonumber\\
    \max\{x_h-a,0\} \quad \text{w.p. } 0.1\nonumber\\
    x_h, \quad \text{otherwise }
    \end{cases}
\end{align}
Thus, if $a=0$, the state $x_{h+1}=x_h$. The agent gets an utility of $g(x_h,a_h,x_{h+1})=(x_{h}-x_{h+1})/2$. We want that utility to be greater than or equal to $4$ at the end of every episode. This will ensure that at most $1$ job can remain at the end of each episode.  

\begin{figure*}
\begin{minipage}{0.48\textwidth}
    \centering
    \includegraphics[clip, trim=0in 4in 0.5in 2in, width=60mm]{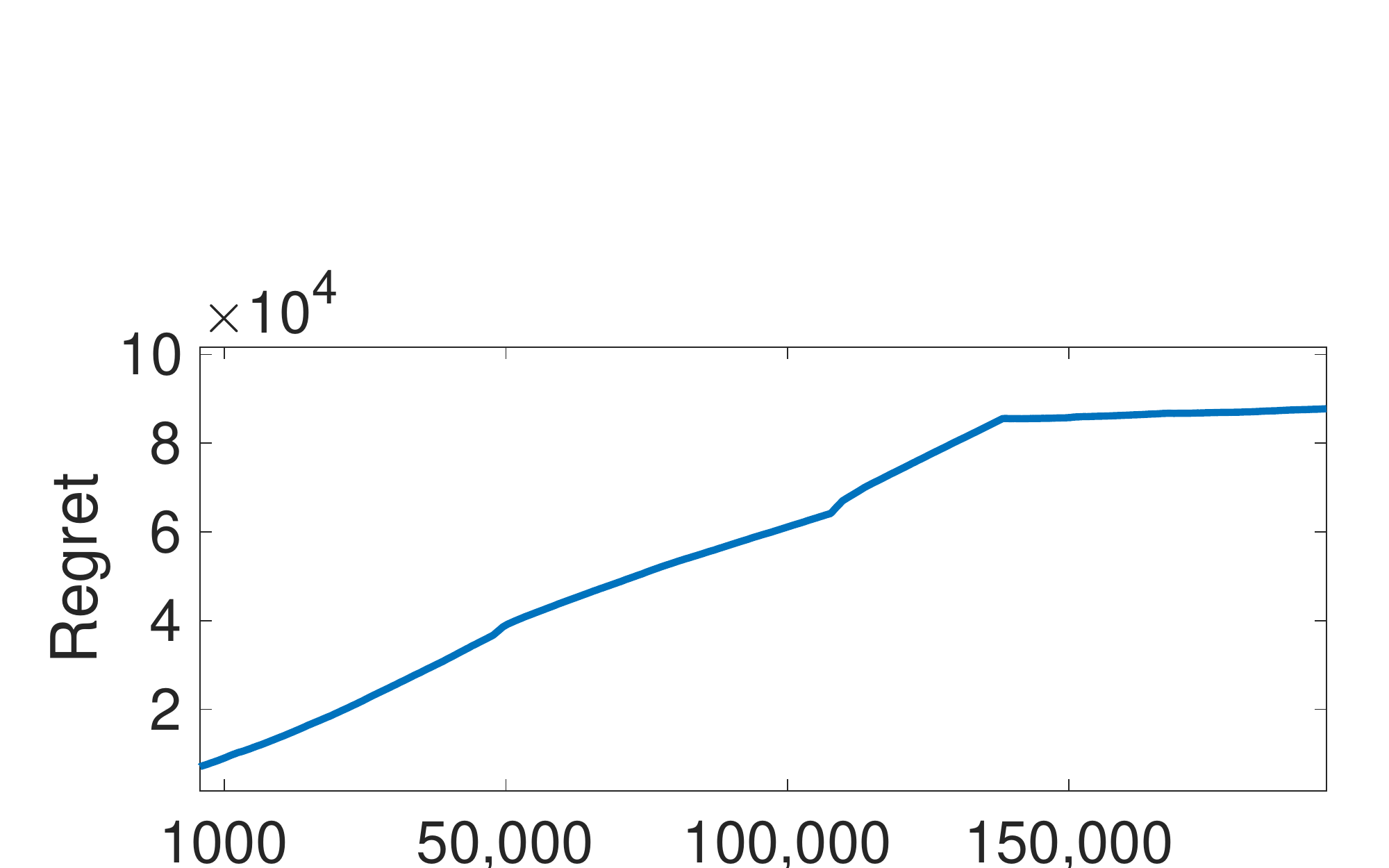}
    \end{minipage}\hfill
    \begin{minipage}{0.48\textwidth}
    \centering
    \includegraphics[clip, trim=0in 4in 0.5in 2in, width=60mm]{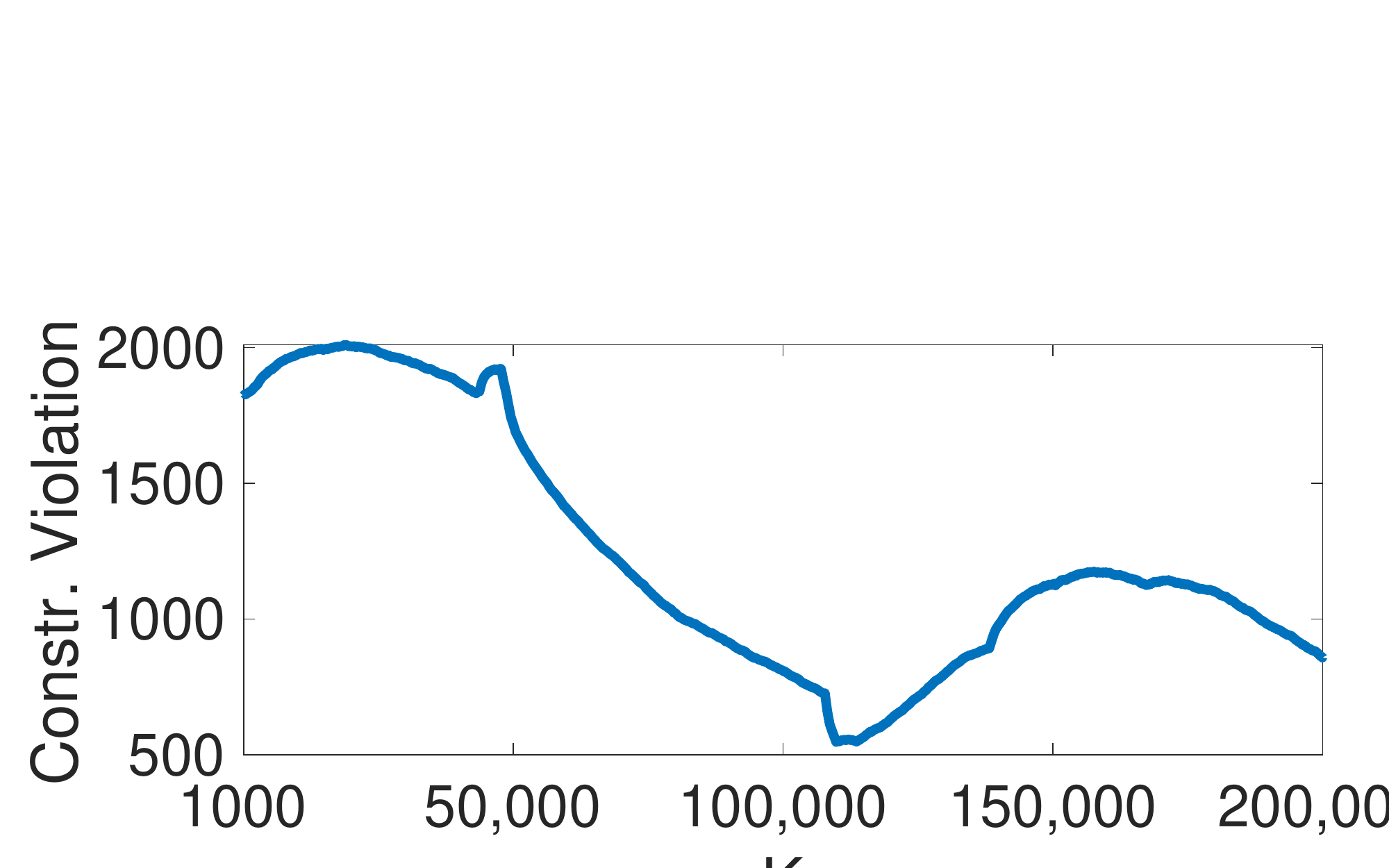}
    \end{minipage}
    \caption{The plot for cumulative regret and constraint violation as a function of $K$. Each plot is an average of $10$ trials.}
    \label{fig:1}
    \vspace{-0.2in}
\end{figure*}

We run Algorithm~\ref{algo:model_free} for $2\times 10^5$ episodes ($K$). The parameters we used are the followings: $\alpha=K/(1+2H/\gamma+H)$,  $\eta=2H/(\gamma\sqrt{KH^2})$. We set $\gamma=1$. We have also set $\epsilon=0.1$ in order to ensure that the violation goes towards $0$ as $K$ increases. Note that the setup can be represented in a tabular form. Since linear MDP contains tabular form, the feature space representation becomes simple, in particular, $\phi(s,a)=\mathbf{e}_{s,a}$ where $\mathbf{e}_{s,a}$ is $1$ for the state-action pair $(s,a)$, and $0$ otherwise. The dimension of the feature space is $|\cS||\cA|$. The cumulative regret and constraint violations are shown in Figure~\ref{fig:1}. As predicted by our theory,  the regret scales only as $\sqrt{K}$. On the other hand, the cumulative violation oscillates. However, the violation approaches $0$ as $K$ increases.
The oscillation of violation is due to the fact that dual variable also oscillates in order to illicit conservative response when the cumulative violation increases, and illicit aggressive response when the violation decreases.  

\section{Conclusion and Future Work}
We propose a model-free RL-based algorithm for linear MDP. We have achieved $\mathcal{\tilde{O}}(\sqrt{d^3H^3T})$ regret and $\mathcal{\tilde{O}}(\sqrt{d^3H^3T})$ constraint violation. To the best of our knowledge, this is the first result which shows $\tilde{\cO}(\sqrt{T})$ regret and $\tilde{\cO}(\sqrt{T})$ constraint violation without requiring a generator for the model-free case. We have extended the LSVI-UCB algorithm in the primal-dual type framework. We have underlined the technical challenges in doing so and explained how the greedy policy fails to achieve an uniform concentration bound for individual value function. Subsequently, we show that a soft-max type algorithm achieves that. 

 Compared to  \cite{ding2021provably}, our regret is off by $\sqrt{d}$ factor which is due to the fact that we need to use uniform concentration bound. The similar gap is also observed for the unconstrained set-up as well. Whether we can tighten this dependence on $d$ remains an important future research direction. Whether we can  tighten the dependence on $H$ also constitutes a future research direction. Extending the work to the setup where the feature-space needs to be learnt is also important. Recent works \cite{modi2021model,zhang2022efficient,agarwal2020flambe} on feature-space learning for unconstrained MDPs may provide some insights in this direction. Consideration of non-linear MDP also constitutes a future research direction. 
 
\textbf{Acknowledgment}

 XZ is partially supported by the NSF grant CNS-2153220. This work has been supported in part by NSF grants: 2112471 (also partly funded by DHS), CNS-2106933, and CNS-1901057 and a grant from the Army Research Office: W911NF-21-1-0244.


\newpage
\clearpage
\bibliographystyle{unsrtnat}
\bibliography{ref}

\newpage
\clearpage

\section*{Checklist}



\begin{enumerate}

\item For all authors...
\begin{enumerate}
  \item Do the main claims made in the abstract and introduction accurately reflect the paper's contributions and scope?
    \answerYes{}
  \item Did you describe the limitations of your work?
    \answerYes{We have clearly compared our results with the existing work. In the Conclusion and Future Work, we identified some important directions where our approach can be extended.} 
  \item Did you discuss any potential negative societal impacts of your work?
    \answerNA{}
  \item Have you read the ethics review guidelines and ensured that your paper conforms to them?
    \answerYes{}
\end{enumerate}

\item If you are including theoretical results...
\begin{enumerate}
  \item Did you state the full set of assumptions of all theoretical results?
    \answerYes{We clearly state the Assumptions 1 and 2 in Section 2}
        \item Did you include complete proofs of all theoretical results?
    \answerYes{We have added the proofs in Appendix.}
\end{enumerate}

\item If you ran experiments...
\begin{enumerate}
  \item Did you include the code, data, and instructions needed to reproduce the main experimental results (either in the supplemental material or as a URL)?
    \answerYes{Our main goal is to obtain Theoretical Results. Nevertheless, we have performed experiment on a simulated setup to verify our theoretical results.}
  \item Did you specify all the training details (e.g., data splits, hyperparameters, how they were chosen)?
    \answerYes{We have characterized the parameters.}
        \item Did you report error bars (e.g., with respect to the random seed after running experiments multiple times)?
    \answerNA{}
        \item Did you include the total amount of compute and the type of resources used (e.g., type of GPUs, internal cluster, or cloud provider)?
    \answerNA{}
\end{enumerate}

\item If you are using existing assets (e.g., code, data, models) or curating/releasing new assets...
\begin{enumerate}
  \item If your work uses existing assets, did you cite the creators?
    \answerNA{}
  \item Did you mention the license of the assets?
    \answerNA{}
  \item Did you include any new assets either in the supplemental material or as a URL?
    \answerNA{}
  \item Did you discuss whether and how consent was obtained from people whose data you're using/curating?
    \answerNA{}
  \item Did you discuss whether the data you are using/curating contains personally identifiable information or offensive content?
    \answerNA{}
\end{enumerate}

\item If you used crowdsourcing or conducted research with human subjects...
\begin{enumerate}
  \item Did you include the full text of instructions given to participants and screenshots, if applicable?
    \answerNA{}
  \item Did you describe any potential participant risks, with links to Institutional Review Board (IRB) approvals, if applicable?
    \answerNA{}
  \item Did you include the estimated hourly wage paid to participants and the total amount spent on participant compensation?
    \answerNA{}
\end{enumerate}

\end{enumerate}
\newpage
\clearpage
\newpage
\clearpage
\appendix
\textbf{Organization of Appendix}: In Section~\ref{sec:preliminary}, we state some results which we use throughout. In Section~\ref{sec:dual_variable}, we prove  Lemma~\ref{lem:dual_variable}. In Section~\ref{sec:proof_step2}, we prove Lemmas~\ref{lem:ucb} and \ref{lem:conc}. In Section~\ref{sec:base_results}, we state and prove base results Lemmas~\ref{lem:phi}, \ref{lem:q_diff}, and \ref{lem:combq_diff} which are necessary to prove the Lemmas~\ref{lem:ucb} and \ref{lem:conc}. Subsequently, we prove Lemmas~\ref{lem:ucb} and \ref{lem:conc} in sequence in Sections~\ref{sec:lem_ucb} and \ref{proof:lem_conc}. In Section \ref{sec:proof_step3}, we prove Theorem~\ref{thm:episodic}. In Section \ref{sec:lem_phi} we prove Lemma~\ref{lem:phi} and highlight that how soft-max policy enables us to obtain a low-covering number for our class of value functions. In Section~\ref{sec:supporting_Results}, we state some results proved in the existing literature which we have used in proving our results. In Section~\ref{sec:nogreedy}, we explain why greedy policy based on the composite $Q$-function fails using an example. In Section~\ref{sec:zero_violation}, we describe the results on how zero violation can be attained. 

\textbf{Notations}: Throughout the rest of this paper, we denote $Q_{r,h}^k, Q_{g,h}^k,w_{r,h}^k, w_{g,h}^k, \Lambda_h^k$ as the $Q$-value and the parameter values estimated at the episode $k$. $V_{j,h}^k(\cdot)=\langle \pi_{h,k}(\cdot|\cdot),Q_{j,h}^{k}(\cdot,\cdot)\rangle_{\cA}$. $\pi_{h,k}(\cdot|x)$ is the soft-max policy based on the composite $Q$-function at the $k$-th episode as $Q_{r,h}^k+Y_kQ_{g,h}^k$. To simplify the presentation, we denote $\phi_h^k=\phi(x_h^k,a_h^k)$. 

Without loss of generality, we assume $||\phi(x,a)||_2\leq 1$ for all $(x,a)\in \cS\times\cA$, $||\mu_h(\cS)||_2\leq \sqrt{d}$, $||\theta_{j,h}||_2\leq \sqrt{d}$ for $j=r,g$ and all $h\in [H]$.

\section{Preliminary Results}\label{sec:preliminary}
\begin{lem}
Under Assumption~\ref{assum:linearmdp}, for any fixed policy $\pi$, let $w_h^{\pi}$ be the corresponding weights such that $Q^{\pi}_{j,h}=\langle \phi(x,a),w_{j,h}^{\pi}\rangle$, for $j\in \{r,g\}$, then we have for all $h\in [H]$, 
\begin{align}
||w_{j,h}^{\pi}||\leq 2H\sqrt{d}
\end{align}
\end{lem}
\begin{proof}
From the linearity of the action-value function, we have
\begin{align}
     Q_{j,h}^{\pi}(x,a)&=j_h(x,a)+\bP_hV_{j,h}^{\pi}(x,a)\nonumber\\
   &  = \langle \phi(x,a),\theta_{j,h}\rangle+\int_{\mathcal{S}}V_{j,h+1}^{\pi}(x^{\prime})\langle \phi(x,a),d\mu_h(x^{\prime})\rangle\nonumber\\
   &  =\langle\phi(x,a),w_{j,h}^{\pi}\rangle
\end{align}
where $w_{j,h}^{\pi}=\theta_{j,h}+\int_{\mathcal{S}}V_{j,h+1}^{\pi}(x^{\prime})d\mu_h(x^{\prime})$.

Now, $||\theta_{j,h}||\leq \sqrt{d}$, and $||\int_{\mathcal{S}}V_{j,h+1}^{\pi}(x^{\prime})d\mu_h(x^{\prime})||\leq H\sqrt{d}$. Thus, the result follows. 
\end{proof}
\begin{lem}\label{lem:w}
For any $(k,h)$, the weight $w_{j,h}^k$ satisfies 
\begin{align}
||w_{j,h}^k||\leq 2H\sqrt{dk/\lambda}
\end{align}
\end{lem}
\begin{proof}
For any vector $v\in \mathcal{R}^d$ we have
\begin{align}\label{eq:inter}
    |v^Tw_{j,h}^k|=|v^T(\Lambda_h^k)^{-1}\sum_{\tau=1}^{k-1}\phi^{\tau}_h(x_h^{\tau},a_h^{\tau})(j_h(x_h^{\tau},a_h^{\tau})+\sum_{a}\pi_{h+1,k}(a|x_{h+1}^{\tau})Q_{j,h+1}^k(x_{h+1}^{\tau},a))|
\end{align}
here $\pi_{h,k}(\cdot|x)$ is the Soft-max policy.

Note that $Q_{j,h+1}^k(x,a)\leq H$ for any $(x,a)$. Hence, from (\ref{eq:inter}) we have
\begin{align}
    |v^Tw_{j,h}^k|& \leq  \sum_{\tau=1}^{k-1}|v^T(\Lambda_h^k)^{-1}\phi^{\tau}_h|.2H\nonumber\\
    & \leq \sqrt{\sum_{\tau=1}^{k-1}v^T(\Lambda^h_k)^{-1}v}\sqrt{\sum_{\tau=1}^{k-1}\phi_h^{\tau}(\Lambda_h^k)^{-1}\phi_h^{\tau}}.2H\nonumber\\
    & \leq 2H||v||\dfrac{\sqrt{dk}}{\sqrt{\lambda}}
\end{align}
Note that $||w_{j,h}^k||=\max_{v:||v||=1}|v^Tw_{j,h}^k|$. Hence, the result follows. 
\end{proof}

\section{Proof of Lemma~\ref{lem:dual_variable}}\label{sec:dual_variable}
We first state and prove the following result which is similar to the one proved in \cite{ding2021provably}. 
\begin{lem}\label{lem:violation}
For $Y\in [0,\xi]$,
\begin{align}
\sum_{k=1}^{K}(Y-Y_k)(b-V_{g,1}^k(x_1))\leq \dfrac{Y^2}{2\eta}+\dfrac{\eta H^2K}{2}
\end{align}
\end{lem}
\begin{proof}
\begin{align}
& |Y_{k+1}-Y|^2=|Proj_{[0,\xi]}(Y_k+\eta(b-V_{g,1}^k(x_1)))-Proj_{[0,\xi]}(Y)|^2\nonumber\\
& \leq (Y_k+\eta(b-V_{g,1}^k(x_1)))-Y)^2\nonumber\\
& \leq (Y_k-Y)^2+\eta^2H^2+2\eta Y_k(b-V_{g,1}^k(x_1))
\end{align}
Summing over $k$, we obtain
\begin{align}
& 0\leq |Y_{K+1}-Y|^2\leq
 |Y_1-Y|^2+2\eta\sum_{k=1}^{K}(b-V_{g,1}^{k}(x_1))(Y_k-Y)+\eta^2H^2K\nonumber\\
& \sum_{k=1}^{K}(Y-Y_k)(b-V_{g,1}^{k}(x_1))\leq
 \dfrac{|Y_1-Y|^2}{2\eta}+\dfrac{\eta H^2K}{2}
\end{align}
Since $Y_1=0$, we have the result. 
\end{proof}
Now, we prove Lemma~\ref{lem:dual_variable}.
\begin{proof}
Note that
\begin{align}
 Y \sum_{k=1}^K (b-V_{g,1}^{\pi_k}(x_1)) & = \sum_{k}(Y-Y_k)(b-V_{g,1}^{k}(x_1))+Y_k(b-V_{g,1}^k)+Y(V_{g,1}^k(x_1)-V_{g,1}^{\pi_k}(x_1))\nonumber\\
  & \leq \dfrac{1}{2\eta}Y^2+\dfrac{\eta}{2}H^2K+\sum_{k=1}^K(Y_kb-Y_kV_{g,1}^k(x_1))+Y(V_{g,1}^k(x_1)-V_{g,1}^{\pi_k}(x_1))\nonumber\\
  & \leq \dfrac{1}{2\eta}Y^2+\dfrac{\eta}{2}H^2K+\sum_{k=1}^K(Y_kV_{g,1}^{\pi^*}(x_1)-Y_kV_{g,1}^k(x_1))+\sum_{k=1}^KY(V_{g,1}^k(x_1)-Y_kV_{g,1}^{\pi_k}(x_1))\nonumber
\end{align}
where the first inequality follows from Lemma~\ref{lem:violation}, and the second inequality follows from the fact that $V_{g,1}^{\pi^*}(x_1)\geq b$. Hence, the result simply follows from the above inequality.
\end{proof}

\section{Proof of Lemmas~\ref{lem:ucb} and \ref{lem:conc}}\label{sec:proof_step2}
First, we prove some base results which we use to prove both Lemmas~\ref{lem:ucb} and \ref{lem:conc} in Section~\ref{sec:base_results}.  Subsequently, we prove Lemma~\ref{lem:ucb} in Section~\ref{sec:lem_ucb} and Lemma~\ref{lem:conc} in Section~\ref{proof:lem_conc}.

\subsection{Proof of Base Results }\label{sec:base_results}
We state and prove  Lemmas~\ref{lem:phi},\ref{lem:q_diff}, and \ref{lem:combq_diff}.

First, we state the concentration lemma which is essential in controlling the fluctuations in the least square value iteration. 

\begin{lem}\label{lem:phi}
There exists a constant $C_2$ such that for any fixed $p\in (0,1)$, if we let $\mathcal{E}$ be the event that
\begin{align}
\norm{\sum_{\tau=1}^{k-1}\phi_{j,h}^{\tau}[V_{j,h+1}^{k}(x_{h+1}^{\tau})-\bP_hV_{j,h+1}^{k}(x_h^{\tau},a_h^{\tau})}_{(\Lambda_h^k)^{-1}}\leq C_2dH\sqrt{\chi}
\end{align}
for all $j\in \{r,g\}$, $\chi=\log[4(C_1+1)\log(|\cA|) dT/p]$, for some constant $C_1$, then $\Pr(\mathcal{E})=1-p/2$. 
\end{lem}
The proof of Lemma~\ref{lem:phi} is technical and relegated to Appendix~\ref{sec:lem_phi}. An extra $\log(|\cA|)$ term appears because $\alpha$ appears in the covering number. 

We now, recursively bound the difference between the value function maintained in Algorithm~\ref{algo:model_free} (without the bonus term) and the value function for any policy for both the reward and utility value functions. We bound this using the expected difference at the next step plus an error term. This error term can be upper bounded by the bonus term with a high-probability.

\begin{lem}\label{lem:q_diff}
There exists an absolute constant $\beta=C_1dH\sqrt{\iota}$, $\iota=\log(\log(|\cA|)4dT/p)$, and for any fixed policy $\pi$, on the event $\mathcal{E}$ defined in Lemma~\ref{lem:phi}, we have 
\begin{align}
\langle \phi(x,a),w_{j,h}^k\rangle-Q_{j,h}^{\pi}(x,a)=\bP_h(V_{j,h+1}^k-V^{\pi}_{j,h+1})(x,a)+\Delta_h^k(x,a)
\end{align}
for some $\Delta_h^k(x,a)$ that satisfies $|\Delta_h^k(x,a)|\leq \beta\sqrt{\phi(x,a)^T(\Lambda_h^k)^{-1}\phi(x,a)}$.
\end{lem}

\begin{proof}
We only prove for $j=r$, the proof for $j=g$ is similar.

Note that $Q_{r,h}^{\pi}(x,a)=\langle\phi(x,a),w_{r,h}^{\pi}\rangle=r_h(x,a)+\bP_hV_{r,h+1}^{\pi}(x,a)$. 


Hence, we have
\begin{align}\label{eq:diff}
& w_{r,h}^k-w_{r,h}^{\pi}=
 (\Lambda_h^k)^{-1}\sum_{\tau=1}^{k-1}\phi_h^{\tau}[r_h^{\tau}+V_{r,h+1}^k(x_{h+1}^{\tau})]
 -w_{r,h}^{\pi}\nonumber\\
& =-\lambda(\Lambda_h^k)^{-1}(w_{r,h}^{\pi})+ (\Lambda_h^k)^{-1}\sum_{\tau=1}^{k-1}\phi_h^{\tau}[V_{r,h+1}^k(x_{h+1}^{\tau})-\bP_hV_{r,h+1}^k(x_{h}^{\tau},a_h^{\tau})]\nonumber\\
& +(\Lambda_h^k)^{-1}\sum_{\tau=1}^{k-1}\phi_h^{\tau}[\bP_hV_{r,h+1}^{k}(x_{h}^{\tau},a_h^{\tau})-\bP_hV_{r,h+1}^{\pi}(x_h^{\tau},a_h^{\tau})]
\end{align}

Now, we bound each term in the right hand side of expression in (\ref{eq:diff}). We call those terms as $\mathbf{q}_1$, $\mathbf{q}_2$, and $\mathbf{q}_3$ respectively. 

First, note that 
\begin{align}\label{eq:diffq1}
|\langle\phi(x,a),\mathbf{q}_1\rangle|& =|\lambda\langle\phi(x,a),( \Lambda_h^k)^{-1}(w_{r,h}^{\pi})\rangle|\nonumber\\
& \leq \sqrt{\lambda}||w_{r,h}^{\pi}||\sqrt{\phi(x,a)^T(\Lambda_h^k)^{-1}\phi(x,a)}
\end{align}
Second, from Lemma~\ref{lem:phi}, for the event in $\mathcal{E}$, we have
\begin{align}\label{eq:diffq2}
|\langle\phi(x,a),\mathbf{q}_2\rangle|\leq CdH\sqrt{\chi}\sqrt{\phi(x,a)^T(\Lambda_h^k)^{-1}\phi(x,a)}
\end{align}
where $\chi=\log(4(C_1+1)\log(|\cA|)dT/p)$.
Third,
\begin{align}\label{eq:diff2}
& \langle \phi(x,a),\mathbf{q}_3\rangle=\langle \phi(x,a),(\Lambda_h^k)^{-1}\sum_{\tau=1}^{k-1}\phi_h^{\tau}[\bP_h(V_{r,h+1}^{k}-V_{r,h+1}^{\pi})(x_h^{\tau},a_h^{\tau})]\rangle\nonumber\\
& =\langle\phi(x,a),(\Lambda_h^k)^{-1}\sum_{\tau=1}^{k-1}\phi_h^{\tau}(\phi_h^{\tau})^T\int(V_{r,h+1}^k-V_{r,h+1}^{\pi})(x^{\prime})d\mu_h(x^{\prime})\rangle\nonumber\\
& =\langle\phi(x,a),\int(V_{r,h+1}^k-V_{r,h+1}^{\pi})(x^{\prime})d\mu_h(x^{\prime})\rangle -\langle\phi(x,a),\lambda(\Lambda_h^k)^{-1}\int(V_{r,h+1}^k-V_{r,h+1}^{\pi})(x^{\prime})d\mu_h(x^{\prime})\rangle
\end{align}
The last term  in (\ref{eq:diff2}) can be bounded as the following
\begin{align}\label{eq:diffq31}
|\langle\phi(x,a),\lambda(\Lambda_h^k)^{-1}\int(V_{r,h+1}^k-V_{r,h+1}^{\pi})(x^{\prime})d\mu_h(x^{\prime})\rangle|\leq 2H\sqrt{d\lambda}\sqrt{\phi(x,a)^T(\Lambda_h^k)^{-1}\phi(x,a)}
\end{align}
since $||\int(V_{r,h+1}^k-V_{r,h+1}^{\pi})(x^{\prime})d\mu_h(x^{\prime})||_2\leq 2H\sqrt{d}$ as $||\mu_h(\cS)||\leq \sqrt{d}$. The first term in (\ref{eq:diff2}) is equal to
\begin{align}\label{eq:diffq32}
\bP_h(V_{r,h+1}^k-V_{r,h+1}^{\pi})(x,a)
\end{align}
Note that $\langle \phi(x,a),w_{r,h}^k\rangle-Q_{r,h}^{\pi}(x,a)=\langle \phi(x,a),w_{r,h}^k-w_{r,h}^{\pi}\rangle=\langle \phi(x,a),\mathbf{q_1}+\mathbf{q_2}+\mathbf{q_3}\rangle$. 
Since $\lambda=1$, we have from (\ref{eq:diffq1}), (\ref{eq:diffq2},(\ref{eq:diffq31}), and (\ref{eq:diffq32})
\begin{align}
    |\langle\phi(x,a),w_{r,h}^k\rangle-Q_{r,h}^{\pi}(x,a)-\bP_h(V_{r,h+1}^k-V_{r,h+1}^{\pi})(x,a)| \leq C_3dH\sqrt{\chi}\sqrt{\phi(x,a)^T(\Lambda_h^k)^{-1}\phi(x,a)}
\end{align}
for some constant $C_3$ which is independent of $C_1$. Finally, note that 
\begin{align}\label{eq:const_beta}
    C_3\sqrt{\chi}&=\sqrt{\log(4(C_1+1)\log(|\cA|)dT/p)}\nonumber\\
    &= C_3\sqrt{\iota+\log(C_1+1)}\nonumber\\
    & \leq C_1\sqrt{\iota}
\end{align}
where $\iota=\log(4\log(|\cA|) dT/p)$. The last inequality follows from the fact that $\iota\in [\log4,\infty)$ as $|A|\geq 2$, and $C_3$ is independent of $C_1$. Hence, we can always pick $C_3\sqrt{\log4+\log(C_1+1)}\leq C_1\sqrt{\log4}$ which satisfies (\ref{eq:const_beta}) for all values of $\iota\in [\log4,\infty)$. 
\end{proof}
Next, using the above lemma, we bound the difference between the composite value function maintained by the algorithm and the composite value function for a policy with the Lagrangian $Y_k$.
 %
 \begin{lem}\label{lem:combq_diff}
 With prob. $1-p$, (for the event in $\cE$)
 \begin{align}\label{eq:q_diff}
 Q_{r,h}^{\pi}(x,a)+Y_kQ_{g,h}^{\pi}(x,a)\leq Q_{r,h}^k(x,a)+Y_kQ_{g,h}^k(x,a)-\bP_h(V_{h+1}^k-V_{h+1}^{\pi,Y_k})(x,a)
 \end{align}
 \end{lem}
 \begin{proof}
 From Lemma~\ref{lem:q_diff} and the fact that $|Q_{r,h}^{\pi}|\leq H$, we have w.p. $1-p/2$,
 \begin{align}
    Q_{r,h}^{\pi}(x,a) &\leq \min\{\langle\phi(x,a),w_{r,h}^k\rangle+\beta\sqrt{\phi(x,a)^T(\Lambda_h^k)^{-1}\phi(x,a)},H\}\nonumber\\
    & +\bP_h(V_{r,h+1}^{\pi}-V_{r,h+1}^{k})(x,a)\nonumber\\
    & = Q_{r,h}^k(x,a)+\bP_h(V_{r,h+1}^{\pi}-V_{r,h+1}^{k})(x,a)\nonumber
 \end{align}
 where the last equality follows from the definition of $Q_{r,h}^k$.
 
 Similarly, with probability $1-p/2$,
 \begin{align}
     Y_kQ_{g,h}^{\pi}(x,a)& \leq Y_kQ_{g,h}^k(x,a)+Y_k\bP_h(V_{g,h+1}^{\pi}-V_{g,h+1}^{k})(x,a)\nonumber
 \end{align}
 Hence, from union bound, with probability $1-p$,
 \begin{align}
     Q_{r,h}^{\pi}(x,a)+Y_kQ_{g,h}^{\pi}(x,a)\leq Q_{r,h}^k+Y_kQ_{g,h}^k(x,a)+\bP_h(V_{h+1}^{\pi,Y_k}-V_{h+1}^k)(x,a)\nonumber
 \end{align}
 \end{proof}

\subsection{Proof of Lemma~\ref{lem:ucb}}\label{sec:lem_ucb}
First, we state and prove a supporting result which bounds the value functions corresponding to the greedy policy and the soft-max policy at a given step. We show that this gap can be controlled by the parameter $\alpha$.

  \begin{lem}\label{lem:close_optimal}
Then, $\bar{V}_{h}^k(x)-V_h^{k}(x)\leq \dfrac{\log|\cA|}{\alpha}$
\end{lem}
where 
\begin{definition}\label{defn:barvhk}
 $\bar{V}_{h}^k(\cdot)=\max_{a}[Q_{r,h}^k(\cdot,a)+Y_kQ_{g,h}^k(\cdot,a)]$.
\end{definition}
$\bar{V}_{h}^k(\cdot)$ is the value function corresponds to the greedy-policy with respect to the composite $Q$-function. 



\begin{proof}
Note that
\begin{align}
V_h^{k}(x)=\sum_{a}\pi_{h,k}(a|x)[Q_{r,h}^k(x,a)+Y_kQ_{g,h}^k(x,a)]
\end{align}
where
\begin{align}\label{eq:boltz}
    \pi_{h,k}(a|x)=\dfrac{\exp(\alpha[Q_{r,h}^k(x,a)+Y_kQ_{g,h}^k(x,a)])}{\sum_{a}\exp(\alpha[Q_{r,h}^k(x,a)+Y_kQ_{g,h}^k(x,a)])}
\end{align}
Denote $a_x=\arg\max_{a}[Q_{r,h}^k(x,a)+Y_kQ_{g,h}^k(x,a)]$

Now, recall from Definition~\ref{defn:barvhk} that $\bar{V}_{h}^k(x)=[Q_{r,h}^k(x,a_x)+Y_kQ_{g,h}^k(x,a_x)]$. Then,
\begin{align}\label{eq:uppb}
    & \bar{V}_{h}^k(x)-V_{h}^{k}(x)=[Q_{r,h}^k(x,a_x)+Y_kQ_{g,h}^k(x,a_x)]\nonumber\\& - \sum_{a}\pi_{h,k}(a|x)[Q_{r,h}^k(x,a)+Y_kQ_{g,h}^k(x,a)]\nonumber\\
    & \leq \left(\dfrac{\log(\sum_{a}\exp(\alpha(Q_{r,h}^k(x,a)+Y_kQ_{g,h}^k(x,a))))}{\alpha}\right)\nonumber\\& -\sum_{a}\pi_{h,k}(a|x)[Q_{r,h}^k(x,a)+Y_kQ_{g,h}^k(x,a)]\nonumber\\
  & \leq \dfrac{\log(|\cA|)}{\alpha}
\end{align}
where the last inequality follows from Proposition 1 in \cite{pan2019reinforcement}.
\end{proof}

We are now ready to show Lemma~\ref{lem:ucb}.
\begin{proof}
We prove the lemma by Induction.


First, we prove for the step $H$. 

 Note that $Q_{j,H+1}^k=0=Q_{j,H+1}^{\pi}$.

Under the event in $\mathcal{E}$ as described in Lemma~\ref{lem:phi} and from Lemma~\ref{lem:q_diff}, we have for $j=r,g$,
\begin{align}
& |\langle\phi(x,a),w_{j,H}^k(x,a)\rangle-Q_{j,H}^{\pi}(x,a)| \leq  \beta\sqrt{\phi(x,a)^T(\Lambda_H^k)^{-1}\phi(x,a)}\nonumber
\end{align}
Hence, for any $(x,a)$,
\begin{align}
 Q_{j,H}^{\pi}(x,a)& \leq \min\{\langle\phi(x,a),w_{j,H}^k\rangle+\beta\sqrt{\phi(x,a)^T(\Lambda_H^k)^{-1}\phi(x,a)},H\}\nonumber\\& 
= Q_{j,H}^k(x,a)
\end{align}
Hence, from the definition of $\bar{V}_h^k$,
\begin{align}
\bar{V}_{H}^k(x)& =\max_{a}[Q_{r,H}^k(x,a)+Y_kQ_{g,h}^k(x,a)]\geq \sum_{a}\pi(a|x)[Q_{r,H}^{\pi}(x,a)+Y_kQ_{g,H}^{\pi}(x,a)] \nonumber\\
& =V_H^{\pi,Y_k}(x)
\end{align}
for any policy $\pi$.  Thus, it also holds for $\pi^{*}$, the optimal policy. Hence, from Lemma~\ref{lem:close_optimal}, we have 
\begin{align}
    V_H^{\pi^*,Y_k}(x)-V_{H}^{k}(x)\leq \dfrac{\log(|\cA|)}{\alpha}\nonumber
\end{align}

Now, suppose that it is true till the step $h+1$ and consider the step $h$.

Since, it is true till step $h+1$, thus, for any policy $\pi$,
\begin{align}
\bP_h(V_{h+1}^{\pi,Y_k}-V_{h+1}^k)(x,a)\leq \dfrac{(H-h)\log(|\cA|)}{\alpha}
\end{align}

From (\ref{eq:q_diff}) in Lemma~\ref{lem:combq_diff} and the above result, we have for any $(x,a)$
\begin{align}
& Q_{r,h}^{\pi}(x,a)+Y_kQ_{g,h}^{\pi}(x,a)  \leq  Q_{r,h}^{k}(x,a)+Y_kQ_{g,h}^k(x,a)+\dfrac{(H-h)\log(|\cA|)}{\alpha}
\end{align}
Hence, 
\begin{align}
    V^{\pi,Y_k}_h(x)\leq \bar{V}_{h}^k(x)+\dfrac{(H-h)\log(|\cA|)}{\alpha}
\end{align}
Now, again from Lemma~\ref{lem:close_optimal}, we have $\bar{V}_{h}^{k}(x)-V_{h}^k(x)\leq \dfrac{\log(|\cA|)}{\alpha}$. Thus,
\begin{align}
    V^{\pi,Y_k}_h(x)-V_{h}^k(x)\leq \dfrac{(H-h+1)\log(|\cA|)}{\alpha}
\end{align}
Now, since it is true for any policy $\pi$, it will be true for $\pi^{*}$. From the definition of $V^{\pi,Y_k}$, we have
\begin{align}
    \left(V^{\pi^*}_{r,h}(x)+Y_kV^{\pi^*}_{g,h}(x)\right)-\left(V_{r,h}^k(x)+Y_kV_{g,h}^k(x)\right)\leq \dfrac{(H-h+1)\log(|\cA|)}{\alpha}
\end{align}
Hence, the result follows by summing over $K$ and considering $h=1$. 
\end{proof}
\subsection{Proof of Lemma~\ref{lem:conc}}\label{proof:lem_conc}
In order to prove the Lemma~\ref{lem:conc}, we state and prove the following result. 
 
 First, we introduce a notation. 
 Let
\begin{align}\label{eq:d_martingale}
    & D_{j,h,1}^k=\langle(Q_{j,h}^k(x_{h}^k,\cdot)-Q_{j,h}^{\pi_{k}}(x_{h}^k,\cdot)),\pi_{h,k}(\cdot|x_h^k)\rangle-(Q_{j,h}^{k}(x_h^k,a_h^k)-Q_{j,h}^{\pi_k}(x_h^k,a_h^k))\nonumber\\
    & D_{j,h,2}^k=\bP_h(V_{j,h+1}^k-V_{j,h+1}^{\pi_k})(x_h^k,a_h^k)-[V_{j,h+1}^k-V_{j,h+1}^{\pi_k}](x_{h+1}^k)
\end{align}

\begin{lem}\label{lm:recursion}
 On the event defined in $\mathcal{E}$ in Lemma~\ref{lem:phi}, we have
 \begin{align}
    V_{j,1}^{k}(x_1)-V_{j,1}^{\pi_k}(x_1)\leq\sum_{h=1}^{H}(D_{j,h,1}^k+D_{j,h,2}^k)+\sum_{h=1}^{H}2\beta\sqrt{\phi(x_{h}^k,a_{h}^k)^T(\Lambda_h^k)^{-1}\phi(x_{h}^k,a_{h}^k)}
\end{align}
\end{lem}
\begin{proof}
By Lemma~\ref{lem:q_diff}, for any $x,h,a,k$
\begin{align}
& \langle w_{j,h}^k(x,a),\phi(x,a)\rangle+\beta\sqrt{\phi(x,a)^T(\Lambda_h^k)^{-1}\phi(x,a)}-Q_{j,h}^{\pi_k} \nonumber\\
& \leq \bP_h(V_{j,h+1}^k-V_{j,h+1}^{\pi_k})(x,a)+2\beta\sqrt{\phi(x,a)^T(\Lambda_h^k)^{-1}\phi(x,a)}
\end{align}
Thus, 
\begin{align}\label{eq:q_d}
Q_{j,h}^{k}(x,a)-Q_{j,h}^{\pi_k}(x,a)\leq \bP_h(V_{j,h+1}^k-V_{j,h+1}^{\pi_k})(x,a)+2\beta\sqrt{\phi(x,a)^T(\Lambda_h^k)^{-1}\phi(x,a)}\nonumber\\
\bP_h(V_{j,h+1}^k-V_{j,h+1}^{\pi_k})(x,a)+2\beta\sqrt{\phi(x,a)^T(\Lambda_h^k)^{-1}\phi(x,a)}-(Q_{j,h}^{k}(x,a)-Q_{j,h}^{\pi_k}(x,a))\geq 0
\end{align}
Since $V_{j,h}^{k}(x)=\sum_{a}\pi_{h,k}(a|x)Q_{j,h}^k(x,a)$ and $V_{j,h}^{\pi_k}(x)=\sum_{a}\pi_{h,k}(a|x)Q_{j,h}^{\pi_k}(x,a)$ where $\pi_{h,k}(a|\cdot)=\textsc{Soft-Max}_{\alpha}^a(Q_{r,h}^k+Y_kQ_{g,h}^k)$ $\forall a$.

Thus, from (\ref{eq:q_d}),
\begin{align}\label{eq:recursive}
    & V_{j,h}^{k}(x_h^k)-V_{j,h}^{\pi_k}(x_h^k)=\sum_{a}\pi_{h,k}(a|x_{h}^k)[Q_{j,h}^{k}(x_{h}^k,a)-Q_{j,h}^{\pi_k}(x_{h}^k,a)]\nonumber\\
    & \leq \sum_{a}\pi_{h,k}(a|x_h^k)[Q_{j,h}^{k}(x_{h}^k,a)-Q_{j,h}^{\pi_k}(x_{h}^k,a)]\nonumber\\
    & +2\beta\sqrt{\phi(x_{h}^k,a_{h}^k)^T(\Lambda_h^k)^{-1}\phi(x_{h}^k,a_{h}^k)}+\bP_h(V_{j,h+1}^k-V_{j,h+1}^{\pi_k})(x_{h}^k,a_h^k)-(Q_{j,h}^{k}(x_h^k,a_h^k)-Q_{j,h}^{\pi_k}(x_h^k,a_h^k))
\end{align}

Thus, from (\ref{eq:recursive}), we have
\begin{align}
    V_{j,h}^k(x_h^k)-V_{j,h}^{\pi_k}(x_h^k)\leq D_{j,h,1}^k+D_{j,h,2}^k+[V_{j,h+1}^k-V_{j,h+1}^{\pi_k}](x_{h+1}^k)+2\beta\sqrt{\phi(x_{h}^k,a_{h}^k)^T(\Lambda_h^k)^{-1}\phi(x_{h}^k,a_{h}^k)}
\end{align}
Hence, by iterating recursively, we have
\begin{align}
    V_{j,1}^{k}(x_1)-V_{j,1}^{\pi_k}(x_1)\leq\sum_{h=1}^{H}(D_{j,h,1}^k+D_{j,h,2}^k)+\sum_{h=1}^{H}2\beta\sqrt{\phi(x_{h}^k,a_{h}^k)^T(\Lambda_h^k)^{-1}\phi(x_{h}^k,a_{h}^k)}
\end{align}
The result follows.
\end{proof}

We, are now ready to prove Lemma~\ref{lem:conc}.
\begin{proof}
Note from Lemma~\ref{lm:recursion}, we have
\begin{align}\label{eq:martin}
\sum_{k=1}^{K}V_{j,1}^{k}(x_1)-V_{j,1}^{\pi_k}(x_1)\leq \sum_{k=1}^{K}\sum_{h=1}^{H}(D_{j,h,1}^k+D_{j,h,2}^k)+\sum_{k=1}^{K}\sum_{h=1}^{H}2\beta\sqrt{\phi(x_h^k,a_h^k)^T(\Lambda_h^k)^{-1}\phi(x_h^k,a_h^k)}
\end{align}
We, now, bound the individual terms. First, we show that the first term corresponds to a Martingale difference.

For any $(k,h)\in [K]\times [H]$, we define $\cF_{h,1}^k$ as $\sigma$-algebra generated by the state-action sequences, reward, and constraint values, $\{(x_i^{\tau},a_i^{\tau})\}_{(\tau,i)\in [k-1]\times [H]}\cup \{(x^k_{i},a_i^k)\}_{i\in [h]}$. 

Similarly, we define the $\cF_{h,2}^k$ as the $\sigma$-algebra generated by $\{(x_i^{\tau},a_i^{\tau})\}_{(\tau,i)\in [k-1]\times [H]}\cup \{(x^k_{i},a_i^k)\}_{i\in [h]}\cup\{x_{h+1}^k\}$. $x_{H+1}^k$ is a null state for any $k\in [K]$. 

A filtration is a sequence of $\sigma$-algebras $\{\cF_{h,m}^k\}_{(k,h,m)\in [K]\times[H]\times[2]}$ in terms of time index
\begin{align}
    t(k,h,m)=2(k-1)H+2(h-1)+m
\end{align}
which holds that $\cF_{h,m}^k\subset \cF_{h^{\prime},m^{\prime}}^{k^{\prime}}$ for any $t\leq t^{\prime}$. 

Note from the definitions in (\ref{eq:d_martingale}) that $D_{j,h,1}^k\in \mathcal{F}_{h,1}^k$ and $D_{j,h,2}^k\in \mathcal{F}_{h,2}^k$. Thus, for any $(k,h)\in [K]\times [H]$, 
\begin{align}
    \mathbbm{E}[D_{j,h,1}^k|\cF_{h-1,2}^k]=0, \quad \mathbbm{E}[D_{j,h,2}^k|\cF_{h,1}^k]=0
\end{align}
Notice that $t(k,0,2)=t(k-1,H,2)=2(H-1)k$. Clearly, $\cF_{0,2}^k=\cF_{H,2}^{k-1}$ for any $k\geq 2$. Let $\cF_{0,2}^1$ be empty. We define a Martingale sequence
\begin{align}
    M_{j,h,m}^k& = \sum_{\tau=1}^{k-1}\sum_{i=1}^{H}(D_{j,i,1}^{\tau}+D_{j,i,2}^{\tau})+\sum_{i=1}^{h-1}(D_{j,i,1}^{k}+D_{j,i,2}^k)+\sum_{l=1}^{m}D_{j,h,l}^k\nonumber\\
    & =\sum_{(\tau,i,l)\in [K]\times[H]\times[2], t(\tau,i,l)\leq t(k,h,m)}D^{\tau}_{j,i,l}
\end{align}
where $t(k,h,m)=2(k-1)H+2(h-1)+m$ is the time index. Clearly, this martingale is adopted to the filtration $\{\cF_{h,m}^k\}_{(k,h,m)\in [K]\times [H]\times [2]}$, and particularly
\begin{align}
    \sum_{k=1}^K\sum_{h=1}^H(D_{j,h,1}^k+D_{j,h,2}^k)=M_{j,H,2}^K
\end{align}

Thus, $M_{j,H,2}^K$ is a Martingale difference satisfying $|M_{j,H,2}^K|\leq 4H$ since $|D_{j,h,1}^k|,|D_{j,h,2}^k|\leq 2H$
From the Azuma-Hoeffding inequality, we have
\begin{align}
\Pr(M_{j,H,2}^K> s)\leq 2\exp(-\dfrac{s^2}{16TH^2})
\end{align}
With probability $1-p/2$ at least for any $j=r,g$,
\begin{align}
\sum_{k}\sum_{h}M_{j,H,2}^K\leq \sqrt{16TH^2\log(4/p)}
\end{align}

Now, we bound the second term. Note that the minimum eigen value of $\Lambda_h^k$ is at least $\lambda=1$ for all $(k,h)\in [K]\times [H]$. By Lemma~\ref{lem:le1}, 
\begin{align}
\sum_{k=1}^{K}(\phi_h^k)^T(\Lambda_h^k)^{-1}\phi_h^k\leq 2\log\left[\dfrac{\det(\Lambda_h^{k+1})}{\det(\Lambda_h^1)}\right]
\end{align}
Moreover, note that $||\Lambda_h^{k+1}||=||\sum_{\tau=1}^{k}\phi_h^k(\phi_h^k)^T+\lambda\bI||\leq \lambda+k$, hence,
\begin{align}
   \sum_{k=1}^{K} (\phi_h^k)^T(\Lambda_h^k)^{-1}\phi_h^k\leq 2d\log\left[\dfrac{\lambda+k}{\lambda}\right]\leq 2d\iota
\end{align}
Now, by Cauchy-Schwartz inequality, we have
\begin{align}
    \sum_{k=1}^{K}\sum_{h=1}^H \sqrt{(\phi_h^k)^T(\Lambda_h^k)^{-1}\phi_h^k}& \leq \sum_{h=1}^{H}\sqrt{K}[\sum_{k=1}^{K}(\phi_h^k)^T(\Lambda_h^k)^{-1}\phi_h^k]^{1/2}\nonumber\\
    & \leq H\sqrt{2dK\iota}
\end{align}
Note that $\beta=C_1dH\sqrt{\iota}$. 

Thus, we have with probability $1-p/2$,
\begin{align}
\sum_{k=1}^{K}V_{r,1}^{k}(x_1)-V_{r,1}^{\pi_k}(x_1)+Y\sum_{k=1}^{K}(V_{g,1}^{k}(x_1)-V_{g,1}^{\pi_k}(x_1))\nonumber\\
\leq  (Y+1)[\sqrt{2TH^2\log(4/p)}+C_4\sqrt{d^3H^3T\iota^2}]
\end{align}
Hence, the result follows. 
\end{proof}

\section{Proof of Theorem~\ref{thm:episodic}}\label{sec:proof_step3}

We, first, show the regret bound.
Note from Lemma~\ref{lem:dual_variable}, Lemma~\ref{lem:ucb}, and Lemma~\ref{lem:conc}, we have for $Y\in [0,\xi]$ at least w.p. $1-p$,
\begin{align}\label{eq:tot}
    \sum_{k=1}^K (V_{r,1}^{\pi^*}(x_1) - V_{r,1}^{\pi_k}(x_1)) + Y \sum_{k=1}^K (b-V_{g,1}^{\pi_k}(x_1))\nonumber\\
    \leq \dfrac{Y^2}{2\eta}+\dfrac{\eta}{2}H^2K+\dfrac{HK\log|\cA|}{\alpha}+\tilde{\cO}((Y+1)\sqrt{d^3H^3T\iota^2})
\end{align}
Replacing $Y$ with $0$ in (\ref{eq:tot}), we have 
\begin{align}
    & \sum_{k=1}^{K}(V_{r,1}^{\pi^*}(x_1)-V_{r,1}^{\pi_k}(x_1)\leq
     \dfrac{\eta}{2}H^2K+\dfrac{HK\log|\cA|}{\alpha}+\mathcal{O}(\sqrt{d^3H^3T\iota^2})
\end{align}
By noting that $\eta=\dfrac{\xi}{\sqrt{KH^2}}$, and $\alpha=\dfrac{\log|\cA|K}{2(1+\xi+H)}$, we have the result. 

We, now, show the violation bound.
Note from Lemma~\ref{lem:dual_variable}, Lemma~\ref{lem:ucb}, and Lemma~\ref{lem:conc} that w.p. $1-p$ (at least),
\begin{align}
& \sum_{k=1}^{K}(V_{r,1}^{\pi^*}(x_1)-V_{r,1}^{\pi_k}(x_1)+Y(b-V_{g,1}^{\pi_k}(x_1))) \leq \dfrac{1}{2\eta}Y^2+\dfrac{\eta}{2}H^2K+\dfrac{HK\log|\cA|}{\alpha}+\nonumber\\& (1+Y)\mathcal{O}(\sqrt{H^3T\iota^2})
\end{align}
Now, put $\eta=\dfrac{\xi}{\sqrt{KH^2}}$, $\alpha=\dfrac{\log|\cA|K}{2(1+\xi+H)}$, and $Y\leq \xi$, then, we have
\begin{align}
   &  \sum_{k=1}^{K}(V_{r,1}^{\pi^*}(x_1)-V_{r,1}^{\pi_k}(x_1)+Y(b-V_{g,1}^{\pi_k}(x_1)))\leq (1+\xi)\mathcal{O}(\sqrt{H^3T\iota^2})+\xi\sqrt{KH^2}
\end{align}
Now, there exists a policy $\pi^{\prime}$ such that $V_{r,1}^{\pi^{\prime}}=\dfrac{1}{K}\sum_{k=1}^{K}V_{r,1}^{\pi_k}$, $V_{g,1}^{\pi^{\prime}}=\dfrac{1}{K}\sum_{k=1}^{K}V_{g,1}^{\pi_k}$. By the occupancy measure, $V_{r,1}^{\pi}$ and $V_{g,1}^{\pi}$ are linear in occupancy measure induced by $\pi$. Thus, the average of $K$ occupancy measure also produces an occupancy measure which induces policy $\pi^{\prime}$ and $V_{r,1}^{\pi^{\prime}}$, and $V_{g,1}^{\pi^{\prime}}$. We take $Y=0$ when $\sum_{k=1}^{K}(b-V_{g,1}^{\pi_k}(x_1^k))<0$, otherwise $Y=\xi$. Hence, we have
\begin{align}
& (V_{r,1}^{\pi^*}(x_1)-\dfrac{1}{K}\sum_{k=1}^{K}V_{r,1}^{\pi_k}(x_1)+\xi(b-\dfrac{1}{K}\sum_{k=1}^{K}V_{g,1}^{\pi_k}(x_1))_{+}\nonumber\\
& =(V_{r,1}^{\pi^*}(x_1)-V_{r,1}^{\pi^{\prime}}(x_1)+\xi[b-V_{g,1}^{\pi^{\prime}}(x_1)]_{+}\nonumber\\
& \leq \dfrac{(1+\xi)\mathcal{O}(\sqrt{d^3H^3T\iota^2})}{K}+\xi\dfrac{\sqrt{KH^2}}{K}\nonumber
\end{align}
Since $\xi=2H/\gamma$, and
using the result of strong duality (Lemma~\ref{lem:convex}), we have
\begin{align}
(b-\dfrac{1}{K}\sum_{k=1}^{K}V_{g,1}^{\pi_k}(x_1^k))_{+}\leq \dfrac{2(1+\xi)}{K\xi}\mathcal{O}(\sqrt{d^3H^3T\iota^2})
\end{align}
Hence, the result follows. 

\section{Proof of Lemma~\ref{lem:phi}}\label{sec:lem_phi}
To simplify the notation, we remove $h$ from the subscript from $Q^k_{j,h}$ and $V^k_{j,h}$ in this Section.

In order to prove the Lemma~\ref{lem:phi}, we first compute the $\epsilon$-covering number for the class of value functions (Lemma~\ref{cor:vcovering}). In order to compute that  we first compute the $\epsilon$-covering number of the individual $Q$-functions (Lemma~\ref{lm:qcovering}) which is essential to compute the covering number for composite $Q$-functions (Corollary~\ref{cor:closeq}). Subsequently, we show that if the two $Q$-functions and the Lagrange multipliers are close, the policies are also close (Lemma~\ref{lem:pi}). 

We first introduce the set of $Q$-functions.
\begin{definition}
Let $\mathcal{Q}_j=\{Q|Q(\cdot,\cdot)=\min\{w_j^T\phi(\cdot,\cdot)+\beta\sqrt{\phi^T(\cdot,\cdot)^T\Lambda^{-1}\phi(\cdot,\cdot)},H\}\}$
\end{definition}
The set $\cQ$ is parameterized by $w_j$,  and $\Lambda$. We have $||w_j||\leq 2H\sqrt{dk/\lambda}$ (from Lemma~\ref{lem:w}). The minimum eigen value of $\Lambda$ satisfies $\lambda_{min}\geq 1$. Hence, the Frobenius norm of $\Lambda^{-1}$ is bounded. Note that $Q_j^k\in \cQ_j$ for $j=r,g$.

We now introduce the class of value function for $j=r,g$.
\begin{definition}\label{defn:classv}
Let $\cV_j=\{V_j|V_j(\cdot)=\sum_{a}\pi(a|\cdot)Q_{j}(\cdot,a); Q_r\in \cQ_r, Q_g\in \cQ_j, Y\in [0,\xi]\}$ for $j=r,g$, where 
\begin{align*}
   \Pi =\{\pi| \forall a\in \cA, \pi(a|\cdot) &= \textsc{Soft-Max}^a_{\alpha}((Q_{r}(\cdot,\cdot)+Y Q_{g}(\cdot,\cdot))
 Q_r \in \cQ_r, Q_g \in \cQ_g, Y \in [0,\xi]\}.
\end{align*}
\end{definition}
The class of value function $\cV_j$ is parameterized by $w_r, w_g$,  $\Lambda$, and $Y\in [0,\xi]$. Note that even the individual value function depends on the $Q$-functions for both the reward and utility since the policy depends on the composite $Q$-function. 

First, we need to see whether $V_j^k\in \cV_j$. Recall the definition of $V_j^k$ at the $k$-th episode $V^k_j(\cdot)=\sum_{a}\pi_k(a|\cdot)Q_{j}^k(\cdot,a)$ where
\begin{equation*}
    \pi_k(a|\cdot)=\textsc{Soft-Max}^a_{\alpha}((Q_{r}(\cdot,\cdot)+Y_kQ_g(\cdot,\cdot)).
\end{equation*}
Since $Q_j\in \cQ_j$ for all $j$, and $0\leq Y_k\leq \xi$, thus, $V_j\in \cV_j$.

We now bound the $\epsilon$-covering number for the class of value function
\begin{lem}\label{cor:vcovering}
     There exists a $\tilde{V}_j\in \mathcal{V}_j$ parameterized by $(\tilde{w}_r,\tilde{w}_g,\tilde{\beta},\Lambda,\tilde{Y})$ such that DIST $(V_j,\tilde{V}_j)\leq \epsilon$ where
    \begin{align}
        \rD\rI\rS\rT(V_j,\tilde{V}_j)=\sup_{x}|V_j(x)-\tilde{V}_r(x)|.
    \end{align}
    Let $N_{\epsilon}^{V_j}$ be the $\epsilon$-covering number for the set $\cV_j$, then, 
    \begin{align}\label{eq:vcovering}
    \log N^{V_j}_{\epsilon}\leq d\log\left(1+8H\dfrac{\sqrt{dk}}{\sqrt{\lambda}\epsilon^{\prime}}\right)+d^2\log\left[1+8d^{1/2}\beta^2/(\lambda(\epsilon^{\prime})^2)\right]+\log\left(1+\dfrac{\xi}{\epsilon^{\prime}}\right)
    \end{align}
    where $\epsilon^{\prime}=\dfrac{\epsilon}{H2\alpha(1+\xi+H)+1}$
\end{lem}
Note that  $\epsilon$-covering number is dependent on $\xi$. This is because the policy depends on the Lagrange multiplier $Y$ which is upper bounded by $\xi$. Thus, we also need $\epsilon$-covering for the Lagrange multiplier in order to obtain $\epsilon$-close value function. Note that the $\epsilon$-covering does not depend on sample dependent terms. Rather it only depends on general $w_{j,h}$, $\Lambda$, and $Y$. Since the policy parameter is $\alpha$, we also have $\epsilon$-covering number is dependent on $\alpha$. 

In order to prove the above lemma, we first state and prove some additional results.

We, first, obtain the $N^{Q_j}_{\epsilon}$ covering number for the set $\cQ_j$. Towards this end, we first, introduce some notations.

\begin{definition}\label{defn:epsilon_cover}
Let $\cC_w^{\epsilon}$ be an $\epsilon/2$- cover of the set $\{w\in \R^d|||w||\leq 2H\sqrt{dk/\lambda}\}$ with respect to the 2-norm. Let $\cC^{\epsilon}_{\vA}$ be an $\epsilon^2/4$-cover of the set $\{\vA\in \R^{d\times d}|||\vA||_F\leq d^{1/2}\beta^2\lambda^{-1}\}$ with respect to the Frobenius norm. 
\end{definition}

\begin{lem}\label{lm:qcovering}
\begin{align}
    |\mathcal{C}_{w}^{\epsilon}|\leq (1+8H\sqrt{dk/\lambda}/\epsilon)^d, \quad |\mathcal{C}_{\vA}^{\epsilon}|\leq [1+8d^{1/2}\beta^2/(\lambda\epsilon^2)]^{d^2}
\end{align}
The $\epsilon$-covering number for the set $\cQ_j$, for $j=r,g$, $N^{Q_j}_{\epsilon}$ of the set  $\mathcal{Q}_j$ for $j=r,g$ satisfies the following
\begin{align}
\log N^{Q_j}_{\epsilon}\leq d\log\left(1+\dfrac{8H\sqrt{dk}}{\sqrt{\lambda}\epsilon}\right)+d^2\log[1+8d^{1/2}\beta^2/(\lambda\epsilon)^2]
\end{align}
 The distance metric is the $\infty$-norm, i.e., $\mathrm{dist}(Q_1,Q_2)=\sup_{x,a}|Q_1(x,a)-Q_2(x,a)|$.
\end{lem}

\begin{proof}
For notational simplicity, we represent $\vA=\beta^2\Lambda^{-1}$, and reparamterized the class $\cQ_j$ by $(w_j,\vA)$.  Now,
\begin{align}
dist(Q_1,Q_2)& = \sup_{x,a}|[w_1^T\phi(x,a)+\sqrt{\phi^T(x,a)\vA_1\phi(x,a)}]-
 [w_2^T\phi(x,a)+\sqrt{\phi^T(x,a)\vA_2\phi(x,a)}]|\nonumber\\
& \leq \sup_{\phi:||\phi||\leq 1}|[w_1^T\phi+\sqrt{\phi^T\vA_1\phi}]-
 [w_2^T\phi+\sqrt{\phi^T\vA_2\phi}]|\nonumber\\
& \leq \sup_{\phi:||\phi||\leq 1}|(w_1-w_2)^T\phi|+\sup_{\phi:||\phi||\leq 1}\sqrt{|\phi^T(\vA_1-\vA_2)\phi|}\nonumber\\
& =||w_1-w_2||+\sqrt{||\vA_1-\vA_2||}\leq ||w_1-w_2||+\sqrt{||\vA_1-\vA_2||_F}
\end{align}
where the second-last inequality follows from the fact that $|\sqrt{x}-\sqrt{y}|\leq \sqrt|x-y|$. For matrices $||\cdot||$, and $||\cdot||_F$ denote matrix operator norm and the Frobenius norm respectively. 

Recall that $\cC_w$ is an $\epsilon/2$- cover of the set $\{w\in \R^d|||w||\leq 2H\sqrt{dk/\lambda}\}$ with respect to the 2-norm. Also recall that $\cC_{\vA}$ be an $\epsilon^2/4$-cover of the set $\{\vA\in \R^{d\times d}|||\vA||_F\leq d^{1/2}\beta^2\lambda^{-1}\}$. Thus, from Lemma~\ref{lem:cover_number},
\begin{align}
|\mathcal{C}_{w}^{\epsilon}|\leq (1+8H\sqrt{dk/\lambda}/\epsilon)^d, \quad |\mathcal{C}_{\vA}^{\epsilon}|\leq [1+8d^{1/2}\beta^2/(\lambda\epsilon^2)]^{d^2}\nonumber
\end{align}
For any $Q_j\in \mathcal{Q}_j$, there exists a $\tilde{Q}_j$ parameterized by $(w_{2},\vA_2)$ where  $w_2\in \cC_w^{\epsilon}$ and $\vA_2\in \cC_{\vA}^{\epsilon}$ such that $\mathrm{dist}(Q_j,\tilde{Q}_j)\leq \epsilon$. Hence, $N^{Q_j}_{\epsilon}\leq |\mathcal{C}^{\epsilon}_{w}||\mathcal{C}^{\epsilon}_{\vA}|$, which gives the result since $\log(\cdot)$ is an increasing function.
\end{proof}
Since the class of $Q$-function is independent of the policy we do not have $\xi$ and $\alpha$ in the $\epsilon$-covering number. 

From the above lemma and since $Y_k\leq \xi$, we have the following,
\begin{cor}\label{cor:closeq}
If $\mathrm{dist}(Q_{r}^k,\tilde{Q}_{r})\leq \epsilon^{\prime}$, $\mathrm{dist}(Q_{g}^k,\tilde{Q}_{g})\leq \epsilon^{\prime}$, and $|\tilde{Y}_k-Y_k|\leq \epsilon^{\prime}$, then, 
$\mathrm{dist}(Q_{r}^k+Y_kQ_{g}^k,\tilde{Q}_r+\tilde{Y}_k\tilde{Q}_g)\leq \epsilon^{\prime}(1+\xi+H)$.  
\end{cor}
\begin{proof} 
Note that $\tilde{Q}_j\in \mathcal{Q}_j$ belongs to the $\epsilon^{\prime}$ covering of the set $\cQ$.
\begin{align}
& \mathrm{dist}(Q_{r}^k+Y_kQ_{g}^k,\tilde{Q}_r+\tilde{Y}_k\tilde{Q}_g)=\sup_{x,a}|(Q_{r}^k(x,a)+Y_kQ_{g}^k(x,a))-(\tilde{Q}_r(x,a)+\tilde{Y}_k\tilde{Q}_g(x,a))|\nonumber\\
& \leq \sup_{x,a}|(Q_{r}^k(x,a)+Y_kQ_{g}^k(x,a))-(\tilde{Q}_{r}(x,a)+Y_k\tilde{Q}_{g}(x,a))|+\sup_{x,a}|(\tilde{Y}_k-Y_k)Q_{g}^k(x,a)|\nonumber\\
& \leq \sup_{x,a}|Q_{r}^k(x,a)-\tilde{Q}_{r}(x,a)|+Y_k\sup_{x,a}|Q_{g}^k(x,a)-\tilde{Q}_{g}(x,a)|+\epsilon^{\prime}H\nonumber\\
& \leq \epsilon^{\prime}(1+Y_k)+\epsilon^{\prime}H\nonumber\\
& \leq \epsilon^{\prime}(1+H+\xi)
\end{align}
where the first inequality follows from the property of supremum and the norm. The second inequality follows from the norm, and the fact that $|\tilde{Y}_k-Y_k|\leq \epsilon^{\prime}$, and $|Q_{g}^k(x,a)|\leq H$. The third inequality follows from the fact that $\mathrm{dist}(Q_j,\tilde{Q}_j)\leq \epsilon^{\prime}$. 
\end{proof}
We now show that if the there exist $\tilde{Q}_j$, and $\tilde{Y}_k$ which are close to $Q_j$ and $Y_k$, then  the soft-max  policy is also close.
\begin{lem}\label{lem:pi}
Suppose that $\pi$ is the soft-max policy (temp. coefficient $1/\alpha$) corresponding to the composite $Q$-functions $(Q_{r}^k+Y_kQ_{g}^k)$, i.e., $\forall a\in \cA$
\begin{align}
    \pi(a|\cdot)=\textsc{Soft-Max}^a_{\alpha}((Q_{r}(\cdot,\cdot)+Y_kQ_g(\cdot,\cdot)).\nonumber
\end{align}
 $\tilde{\pi}$ is the soft-max policy vector with the same temp. coefficient $1/\alpha$ corresponding to the composite $Q$-function $(\tilde{Q}_r+\tilde{Y}_k\tilde{Q}_g)$, i.e, $\forall a\in \cA$,
 \begin{align}
     \tilde{\pi}(a|\cdot)=\textsc{Soft-Max}^a_{\alpha}((\tilde{Q}_{r}(\cdot,\cdot)+\tilde{Y}_k\tilde{Q}_g(\cdot,\cdot)).\nonumber
 \end{align}then, for any state $x$,
\begin{align}
||\pi(\cdot|x)-\tilde{\pi}(\cdot|x)||_1\leq 2\alpha\epsilon^{\prime}(1+\xi+H)
\end{align}
where $\pi(\cdot|x)=\{\pi(a|x)\}_{a\in \cA}$ and $\tilde{\pi}(\cdot|x)=\{\tilde{\pi}(a|x)\}_{a\in \cA}$ 
when $\mathrm{dist}(Q_{j,h}^k,\tilde{Q}_{j})\leq \epsilon^{\prime}$ for $j=r,g$, and $|\tilde{Y}_k-Y_k|\leq \epsilon^{\prime}$. 
\end{lem}


\begin{proof}
Let $\mathrm{Exp}^{\alpha}(P)$ be a soft-max  corresponding to the vector $P$, i.e., the $i$-th component of $\mathrm{Exp}^{\alpha}(P)$ is
\begin{align}
    \dfrac{\exp(\alpha P_i)}{\sum_{i}\exp(\alpha P_i)}.\nonumber
\end{align}
Note from Theorem 4.4 in \cite{epasto2020optimal} then, we have
\begin{align}\label{eq:use_result}
    ||\mathrm{Exp}^{\alpha}(P_1)-\mathrm{Exp}^{\alpha}(P_2)||_1\leq 2\alpha||P_1-P_2||_{\infty}
\end{align}
for two vectors $P_1$ and $P_2$. 

Now note that in our case for a given state $x$, $\pi$ is equivalent to $\mathrm{Exp}^{\alpha}(Q_{r,h}^k(x,\cdot)+Y_kQ_{g,h}^k(x,\cdot))$, and $\tilde{\pi}$ is equivalent to $\mathrm{Exp}^{\alpha}(\tilde{Q}_r(x,\cdot)+\tilde{Y}_k\tilde{Q}_g(x,\cdot))$. Then from (\ref{eq:use_result}) and the fact that $\mathrm{dist}(Q_{r,h}^k+Y_kQ_{g,h}^k,\tilde{Q}_r+\tilde{Y}_k\tilde{Q}_g)\leq \epsilon^{\prime}(1+\xi+H)$ (by Corollary~\ref{cor:closeq}) we have 
\begin{align}
    ||\pi(\cdot|x)-\tilde{\pi}(\cdot|x)||_1\leq 2\alpha\epsilon^{\prime}(1+\xi+H)
\end{align}
Hence, the result follows.
\end{proof}

Based on the above two lemmas we show that when the $Q$-functions are close, the value functions in the class $\cV_j$ are also close. 


\begin{lem}\label{lem:vclose}
There exists $\tilde{V}_j\in \cV_j$ such that
\begin{align}\label{eq:vclose}
\mathrm{DIST}(V_{j}^k,\widetilde{V}_{j})\leq H2\alpha\epsilon^{\prime}(1+\xi+H)+\epsilon^{\prime},
\end{align} 
where  $\mathrm{dist}(\tilde{Q}_j,Q_j)\leq\epsilon^{\prime}$, $\tilde{Q}_j\in\cQ_j$ for all $j$;
\begin{align}
    \widetilde{V}_j(\cdot)=\sum_{a}[\tilde{\pi}(a|\cdot)\tilde{Q}_j(\cdot)],\nonumber
\end{align}
\begin{align}
    \tilde{\pi}(a|\cdot)=\textsc{Soft-Max}^a_{\alpha}((\tilde{Q}_{r}(\cdot,\cdot)+\tilde{Y}_k\tilde{Q}_g(\cdot,\cdot)),\quad  \forall a\in \cA\nonumber
\end{align}
 $|\tilde{Y}_k-Y_k|\leq \epsilon^{\prime}$.
\end{lem}
\begin{proof}
For any $x$,
\begin{align}
& V_{j}^k(x)-\widetilde{V}_j(x)\nonumber\\
    & = |\sum_{a}\pi(a|x)Q_{j}^k(x,a)-\sum_{a}\tilde{\pi}(a|x)\tilde{Q}_{j}(x,a)|\nonumber\\
    & =|\sum_{a}\pi(a|x)Q_{j}^k(x,a)-\sum_{a}\pi(a|x)\tilde{Q}_{j}(x,a)+\sum_{a}\pi(a|x)\tilde{Q}_j(x,a)-\sum_{a}\tilde{\pi}(a|x)\tilde{Q}_j(x,a)|\nonumber\\
    & \leq |\sum_{a}\pi(a|x)Q_{j}^k(x,a)-\sum_{a}\pi(a|x)\tilde{Q}_{j}(x,a)|+|\sum_{a}\pi(a|x)\tilde{Q}_j(x,a)-\sum_{a}\tilde{\pi}(a|x)\tilde{Q}_j(x,a)|\nonumber\\
    & \leq \epsilon^{\prime}+||\pi(\cdot|x)-\tilde{\pi}(\cdot|x)||_1||\tilde{Q}_j(x)||_{\infty}\nonumber\\
    & \leq \epsilon^{\prime}+H2\alpha\epsilon^{\prime}(1+\xi+H)
\end{align}
where we use the fact that $\mathrm{dist}(Q_{j}^k,\tilde{Q}_r)\leq \epsilon^{\prime}$, and $\sum_{a}\pi(a|x)= 1$ for the first term and the Holder's inequality in the second  term for the second last inequality. For the last inequality, we use Lemma~\ref{lem:pi}, and the fact that $\tilde{Q}_j(x,a)\leq H$ for any $(x,a)$.
 Hence, we have the result.
\end{proof}
 Note that when $\alpha=\dfrac{\log(|\cA|)K}{2(1+\xi+H)}$ as we have in Algorithm~\ref{algo:model_free}, the right hand side in (\ref{eq:vclose}) becomes
 \begin{align}
     \epsilon^{\prime}+\log(|\cA|)KH\epsilon^{\prime}
 \end{align}

We introduce one more notation which we use to prove Lemma~\ref{cor:vcovering}.
\begin{definition}\label{defn:xicovering}
Let $\cC_{\xi}^{\epsilon}$ be an $\epsilon$ cover for $Y\in [0,\xi]$. Hence,  $|\cC_{\xi}^{\epsilon}|\leq \left(1+\dfrac{\xi}{\epsilon}\right)$
\end{definition}
Note that $\cC_{\xi}^{\epsilon}$ consists of points which is $\epsilon$-close to any point within the interval $[0,\xi]$. Since we have defined $\epsilon$-cover for all the parameters, we  are now ready to prove Lemma~\ref{cor:vcovering}.

\begin{proof}
Fix an $\epsilon$.
Let $\epsilon^{\prime}=\dfrac{\epsilon}{H2\alpha(1+\xi+H)+1}$, then from Lemma~\ref{lem:vclose},  we have DIST$(V_{j}^k,\widetilde{V}_j)\leq\epsilon$. Thus, we only need to find parameters in the  $\epsilon^{\prime}$-covering of the $Q$-functions as described in Lemma~\ref{lm:qcovering} in order to obtain $\epsilon$-close value function.

Recall the Definition~\ref{defn:epsilon_cover}. Then, there exists $\tilde{w}_r,\tilde{w}_g\in \cC_w^{\epsilon^{\prime}}$ such that
$||\tilde{w}_r-w_r||\leq \dfrac{\epsilon^{\prime}}{2}$, $||\tilde{w}_g-w_g||\leq \dfrac{\epsilon^{\prime}}{2}$. Further, there exists $\vA_2\in \cC_{\vA}^{\epsilon^{\prime}}$   such that $||\vA-\tilde{\vA}||_F\leq \dfrac{\epsilon^{\prime 2}}{4}$, $\vA=\beta^2(\Lambda^k)^{-1}$, $\tilde{\vA}=\beta^2(\tilde{\Lambda})^{-1}$, for some $\tilde{\Lambda}$, and $Y_k,\tilde{Y}_k$ such that $|Y_k-\tilde{Y}_k|\leq \epsilon^{\prime}$. Then we obtain $\tilde{Q}_j$ parameterized by $(\tilde{w}_j,\beta,\tilde{\Lambda})$ for $j=r,g$, such that $\mathrm{dist}(Q_j,\tilde{Q}_j)\leq \epsilon^{\prime}$ (by Lemma~\ref{lm:qcovering}).

Now define $\widetilde{V}_j=\sum_{a}\tilde{\pi}(a|\cdot)\tilde{Q}_j$, where 
\begin{align}
    \tilde{\pi}(a|\cdot)=\textsc{Soft-Max}^a_{\alpha}((\tilde{Q}_{r}(\cdot,\cdot)+\tilde{Y}_k\tilde{Q}_g(\cdot,\cdot)).\nonumber
\end{align}
  Thus, from Lemma~\ref{lem:vclose}, we have DIST$(V_{j}^k,\tilde{V}_j)\leq\epsilon$. Hence, there exists $\tilde{V}_j$ parameterized by $\tilde{w}_r,\tilde{w}_g,\tilde{Y}_k,\tilde{\vA}$, such that Dist($\tilde{V}_j,V_j^k)\leq \epsilon$. Hence, $N_{\epsilon}^V\leq |\cC_w^{\epsilon^{\prime}}||\cC_{\vA}^{\epsilon^{\prime}}||\cC_{\xi}^{\epsilon^{\prime}}|$. Thus, from Lemma~\ref{lm:qcovering} and Definition~\ref{defn:xicovering},  the $\epsilon$-covering number $N_{\epsilon}^{V_j}$ for the set $\cV_j$ satisfies the following
\begin{align}
    \log N^{V_j}_{\epsilon}\leq d\log\left(1+8H\dfrac{\sqrt{dk}}{\sqrt{\lambda}\epsilon^{\prime}}\right)+d^2\log\left[1+8d^{1/2}\beta^2/(\lambda(\epsilon^{\prime})^2)\right]+\log\left(1+\dfrac{\xi}{\epsilon^{\prime}}\right). \nonumber
\end{align}

Hence, the result follows.
\end{proof}

From Lemma~\ref{cor:vcovering}, note that we need $\epsilon^{\prime}$ covering for the $Q$-functions where $\epsilon^{\prime}=\dfrac{\epsilon}{(H2\alpha(1+\xi+H)+1)}$ if we need to bound DIST $(V_j,\tilde{V}_j)$ by $\epsilon$. 

Now, we are ready to prove Lemma~\ref{lem:phi}. 

\begin{proof}
By Lemma~\ref{cor:vcovering}, we know that there exists $\tilde{V}_j$ in the $\epsilon$-covering for  $\cV_j$ such that for every $x$,
\begin{align}\label{eq:delv}
    V_j(x)=\tilde{V}_j(x)+\Delta V(x)
\end{align}
where $\sup_{x}\Delta V(x)\leq \epsilon$. 

Hence, 
\begin{align}
    \norm{\sum_{\tau=1}^{k}\phi^{\tau}(V_j(x_{\tau})-\mathbbm{E}[V_j(x_{\tau})|\mathcal{F}_{\tau-1}])}^2_{(\Lambda^k)^{-1}}& \leq  
2\norm{\sum_{\tau=1}^{k}\phi^{\tau}(\tilde{V}_j(x_{\tau})-\mathbbm{E}[\tilde{V}_j(x_{\tau})|\mathcal{F}_{\tau-1}])}^2_{(\Lambda^k)^{-1}}\nonumber\\& +
2\norm{\sum_{\tau=1}^{k}\phi^{\tau}(\Delta V(x_{\tau})-\mathbbm{E}[\Delta V(x_{\tau})|\mathcal{F}_{\tau-1}])}^2_{(\Lambda^k)^{-1}}
\end{align}
The last expression is bounded by $\dfrac{8k^2\epsilon^2}{\lambda}$. 

Now, we bound the first term. Note from Lemma~\ref{cor:vcovering} that in order to obtain $\tilde{V}_j$ which satisfies (\ref{eq:delv}), we need to obtain  we need $N^V_{\epsilon}$ number of elements to obtain such $(\tilde{w}_r,\tilde{w}_g,\beta,\tilde{\Lambda},\tilde{Y})$.  Such $\tilde{V}_j$ is independent of samples. Hence, we can use the Elliptical lemma for self-normalization (Theorem~\ref{thm:self_norm}). From Theorem~\ref{thm:self_norm} and the union bound we obtain
\begin{align}\label{eq:inequal}
    \norm{\sum_{\tau=1}^{k}\phi^{\tau}(\tilde{V}_j(x_{\tau})-\mathbbm{E}[\tilde{V}_j(x_{\tau})|\mathcal{F}_{\tau-1}])}^2_{(\Lambda^k)^{-1}}\leq 2H^2\left[d\log\left(\dfrac{k+\lambda}{\lambda}\right)+\log\left(\dfrac{N_{\epsilon}^V}{\delta}\right)\right]
\end{align}
where $N^V_{\epsilon}$ is upper bounded in (\ref{eq:vcovering}).   $\beta$ is equal to $C_1dH\sqrt{\iota}$ for some constant $C_1$, and $\iota=\log(\log(|\cA|)4dT/p)$.  Further, $\xi=2H/\gamma$ (by Definition~\ref{defn:xi}). We obtain from (\ref{eq:inequal})
\begin{align}\label{eq:inequal2}
& \norm{\sum_{\tau=1}^{k}\phi^{\tau}(\tilde{V}_j(x_{\tau})-\mathbbm{E}[\tilde{V}_j(x_{\tau})|\mathcal{F}_{\tau-1}])}^2_{(\Lambda^k)^{-1}}\leq\nonumber\\ &
    4H^2\left[\dfrac{d}{2}\log\left(\dfrac{k+\lambda}{\lambda}\right)+d\log\left(1+\dfrac{8H\sqrt{dk}}{\epsilon^{\prime}\sqrt{\lambda}}\right)+d^2\log\left(1+\dfrac{8d^{1/2}\beta^2}{\epsilon^{\prime2}\lambda}\right)+\log\left(1+\dfrac{2H}{\gamma\epsilon^{\prime}}\right)+\log\left(\dfrac{4}{p}\right)\right]
\end{align}
where $\epsilon^{\prime}=\dfrac{\epsilon}{(H2\alpha(1+\xi+H)+1)}$.
Set $\epsilon=\dfrac{dH}{k}$ 
, $\lambda=1$. Thus, $\epsilon^{\prime}=\dfrac{dH}{(2H\alpha(1+\xi+H)+1)k}$. Plugging in the above,  and putting $\alpha=\dfrac{\log(|\cA|)K}{2(1+\xi+H)}$, we obtain from (\ref{eq:inequal2})
\begin{align}
& ||\sum_{\tau=1}^{k}\phi^{\tau}(\tilde{V}_j(x_{\tau})-\mathbbm{E}[\tilde{V}_j(x_{\tau})|\mathcal{F}_{\tau-1}])||^2_{\Lambda_k^{-1}}\leq
C_2H^2d^2\log\left(\dfrac{4(C_1+1)\log(|\cA|)dT}{p}\right)
\end{align}
 for some constant $C_2$, and $C_1=256(1+1/\gamma)$. Hence, the result follows. 

\end{proof}
\section{Supporting Results}\label{sec:supporting_Results}
The following result is shown in \cite{abbasi2011improved} and in Lemma D.2 in \cite{jin2020provably}.
\begin{lem}\label{lem:le1}
Let $\{\phi_t\}_{t\geq 0}$ be a sequence in $\Re^d$ satisfying $\sup_{t\geq 0}||\phi_t||\leq 1$. For any $t\geq 0$, we define $\Lambda_t=\Lambda_0+\sum_{j=0}^{t}\phi_j\phi_j^T\phi_j$. Then if the smallest eigen value of $\Lambda_0$ be at least $1$, we have
\begin{align}
   \log\left[\dfrac{\det(\Lambda_h^{k+1})}{\det(\Lambda_h^1)}\right]\leq \sum_{k=1}^{K}(\phi_h^k)^T(\Lambda_h^k)^{-1}\phi_h^k\leq 2\log\left[\dfrac{\det(\Lambda_h^{k+1})}{\det(\Lambda_h^1)}\right]
\end{align}
\end{lem}
\begin{theorem}\label{thm:self_norm}[Concentration of Self-Normalized Process \cite{abbasi2011improved}]
Let $\{\epsilon_t\}_{t=1}^{\infty}$ be a real-valued stochastic process with corresponding filtration $\{\cF_t\}_{t=0}^{\infty}$. Let $\epsilon_t|\cF_{t-1}$ be a zero mean and $\sigma$ sub-Gaussian, i.e., $\bE[\epsilon_t|\cF_{t-1}]=0$, and
\begin{align}
    \forall \zeta\in \Re, \quad \bE[e^{\zeta\epsilon_t}|\cF_{t-1}]\leq e^{\zeta^2\sigma^2/2}.
\end{align}
Let $\{\phi_t\}_{t=1}^{\infty}$ be a $\Re^d$-valued Stochastic process where $\phi_t\in \cF_{t-1} $. Assume $\Lambda_0\in \Re^{d\times d}$ is a positive-define matrix,  let,  $\Lambda_t=\Lambda_0+\sum_{j=0}^{t}\phi_j\phi_j^T\phi_j$. Then for any $\delta>0$ with probability at least $1-\delta$, we have 
\begin{align}
    ||\sum_{s=1}^{t}\phi_s\epsilon_s||_{\Lambda_t^{-1}}^2\leq 2\sigma^2\log\left[\dfrac{\det(\Lambda_t)^{1/2}\det(\Lambda_0)^{-1/2}}{\delta} \right]
\end{align}
\end{theorem}
The next result characterizes the covering number of an Euclidean ball (Lemma 5.2 in \cite{vershynin2010introduction}). 
\begin{lem}\label{lem:cover_number}[Covering Number of Euclidean Ball]
For any $\epsilon>0$, the $\epsilon$-covering number of the Euclidean ball in $\R^d$ with radius $R$ is upper bounded by $(1+2R/\epsilon)^d$.
\end{lem}
We have used the following result from the optimization which is proved in Lemma 9 in \cite{ding2021provably}.
\begin{lem}\label{lem:convex}
Let $Y^*$ be the optimal dual variable, and $C\geq 2Y^*$, then, if 
\begin{align}
V_{r,1}^{\pi^*}(x_1)-V_{r,1}^{\tilde{\pi}}(x_1)+C[b-V_{g,1}^{\tilde{\pi}}(x_1)]_{+}\leq \delta
\end{align}
then
\begin{align}
[b-V_{g,1}^{\tilde{\pi}}(x_1)]_{+}\leq \dfrac{2\delta}{C}.
\end{align}
\end{lem}
\section{Why does the Greedy Policy Fail?}
\label{sec:nogreedy}
In this section, using an example we show that the greedy-policy is not Lipschitz. Further, we can not use the greedy-policy on the composite $Q$-function to obtain the $\epsilon$-close covering for individual reward and utility value functions. 

Consider the following toy-example: Suppose that the cardinality of the action space $|A|$ is $2$. $Q_{r,h}(x,a_1)=M$, $Q_{r,h}(x,a_2)=1$, $Q_{g,h}(x,a_1)=1$, $Q_{g,h}(x,a_2)=M+\epsilon/2$. Consider $Y=1$, then, greedy policy based on the composite $Q$-function is to choose action $a_2$. 

Note the $\epsilon$-closest values can be anywhere in the ball within $\epsilon$ radius centered around $Q_{j,h}$, and $Y$. Assume that the closest $\epsilon$-cover for $Q_{j,h}$ be $\tilde{Q}_{j,h}$ such that  $\tilde{Q}_{r,h}(x,a_1)= M+\epsilon/2$, $\tilde{Q}_{r,h}(x,a_2)=1-\epsilon/2$, $\tilde{Q}_{g,h}(x,a_1)=1+\epsilon/2$, $\tilde{Q}_{g,h}(x,a_2)=M$, and $\tilde{Y}=1-\epsilon/2$. Then, we have the composite $Q$-functions as
\begin{align}
    & \tilde{Q}_{r,h}(x,a_1)+\tilde{Y}\tilde{Q}_{g,h}(x,a_1)=M+\epsilon/2+(1-\epsilon/2)(1+\epsilon/2)\nonumber\\
   &  \tilde{Q}_{r,h}(x,a_2)+\tilde{Y}\tilde{Q}_{g,h}(x,a_2)=1-\epsilon/2+(1-\epsilon/2)M
\end{align}Hence, it is clear that  the greedy policy based on the composite $Q$-function  is to choose action $a_1$. Hence, the policy is not Lipschitz. Thus, even though the change in the $Q$-function is only by $\epsilon$-amount the decision changes from taking action $a_2$ to taking action $a_1$ in a deterministic fashion.


Now, we see the changes in the value function. Since the policy is greedy and the policy is to choose $a_2$ for the $Q$-function $Q_{j,h}$, then,  $V_{r,h}(x)=Q_{r,h}(x,a_2)=1$. On the other hand, the policy is to choose $a_1$ for the $Q$-functions $\tilde{Q}_{j,h}$. Hence,  $\tilde{V}_{r,h}(x)=M+\epsilon/2$, hence, $|V^{k}_{r,h}(x)-\tilde{V}_{r,h}(x)|>M-1$, and can be made arbitrarily large by making $M$ arbitrarily large. Thus, the individual value functions can not be made close even  though $Q_{j,h}$ and $\tilde{Q}_{j,h}$ are close. This is the reason we can not obtain $\epsilon$-covering number if the greedy policy is based on the composite $Q$-functions. 

Note that in the unconstrained case (equivalent to $Y_k=0$), the decision would be to choose $a_1$ for both $Q_{r,h}$ and $\tilde{Q}_{r,h}$ if the policy is set at the greedy one. Hence, the value function would only differ by at most $\epsilon$-amount. Hence, the greedy policy works for the unconstrained case.

\section{Analysis for Zero Constraint Violation}\label{sec:zero_violation}

The main idea behind attaining zero constraint violation is to consider the following tighter optimization problem--
\begin{align}\label{eq:cmdp_close}
  \text{maximize }_{\pi\in \Delta(\mathcal{A}|\mathcal{S},\mathcal{H})} V^{\pi}_{r,1}(x_1)
\quad \text{subject to } V^{\pi}_{g,1}(x_1)\geq b+\zeta.
 \end{align}
 Since we replace $b$ by $b+\zeta$, we are basically solving the above tighter optimization problem. By ensuring that $\zeta\leq \gamma/2$, we can ensure that Slater's condition is always satisfied, and strong duality holds.  We show that by carefully choosing $\zeta $, we can achieve zero constraint violation with the same order on regret with respect to $T$. First, we introduce some notations which we use throughout this section.

 Let $\pi^{\zeta,*}$ be the optimal solution of the optimization problem in (\ref{eq:cmdp_close}). Since the Slater's condition holds, the strong duality holds by \cite{paternain2019safe}. The optimal dual variable $Y^{\zeta}$ of this tighter problem is 
 \begin{align}
     Y^{\zeta}\leq \dfrac{V^{\pi^{\zeta,*}}(x_1)-V^{\bar{\pi}}_{r,1}(x_1)}{b+\gamma-(b+\zeta)}\leq 4H/\gamma
 \end{align}
 where the last inequality follows from the fact that $\zeta\leq \gamma/2$.
 
 Now, we state the main result---
 \begin{theorem}\label{thm:zero_violation}
    In Algorithm~\ref{algo:model_free}, replacing $b=b+\zeta$, and $\xi=4H/\gamma$. We obtain, with probability $1-p$,
    \begin{align}
        & \mathrm{Regret} (K)\leq C\cO(\sqrt{d^3H^3T\iota^2})+KH\zeta/\delta\nonumber\\
        & \mathrm{Violation} (K)\leq \max\{C^{\prime}\cO(\dfrac{2(1+\xi)}{\xi}\sqrt{d^3H^3T\iota^2})-K\zeta,0\}
    \end{align}
    where $\zeta=\min\{C^{\prime}\cO\left(\dfrac{2(1+\xi)}{\xi}\dfrac{\sqrt{d^3H^3T\iota^2}}{K}\right),\gamma/2\}$.
\end{theorem}
When $C^{\prime}\cO\left(\dfrac{2(1+\xi)}{\xi}\dfrac{\sqrt{d^3H^3T\iota^2}}{K}\right)\leq \gamma/2
$, then the constraint is upper bounded by $0$. Hence, for large enough $K$, we can achieve zero violation. However, by plugging the value of $\zeta$, we obtain the upper bound on regret as
\begin{align}
\mathrm{Regret}(K)\leq C\cO(\sqrt{d^3H^3T\iota^2})+C^{\prime}H\cO\left(\dfrac{2(1+\xi)}{\xi}\sqrt{d^3H^3T\iota^2}\right)\nonumber
\end{align}
where we replace the upper bound of $\zeta$ by $C^{\prime}\cO\left(\dfrac{2(1+\xi)}{\xi}\dfrac{\sqrt{d^3H^3T\iota^2}}{K}\right)$. Thus, the upper bound on regret is
$\tilde{\cO}(\sqrt{d^3H^5T})$. Hence, the order on regret with respect to $T$ is maintained. However, there is an additional $H$ factor in front of the regret. A concurrent work \cite{vaswani2022near} on model-based discounted horizon  tabular setup using a generator model shows that extra $H$ is unavoidable if one wants to achieve zero violation.\footnote{\cite{vaswani2022near} provides a sample complexity guarantee for the discounted horizon setup with discount factor $\gamma$. One can convert the result in the discounted setup to the episodic setup by equating $1/(1-\gamma)=H$.} Even though the setup is different, it seems that the extra $H$ factor is unavoidable.

\begin{proof}
First, we prove the upper bound on regret.
The regret can be decomposed as the following -
\begin{align}\label{eq:tighter_regret}
    \mathrm{Regret}(K)=\sum_{k=1}^{K}(V^{\pi^*}_{r,1}(x_1)-V^{\pi^{\zeta,*}}(x_1))+\sum_{k=1}^{K}(V_{r,1}^{\pi^{\zeta,*}}(x_1)-V^{\pi_k}(x_1))
\end{align}
The first term can be bounded with the help of the following lemma (the proof for finite state is in \cite{wei2021provably}, the extension to linear MDP is provided after this proof)--
\begin{lem}\label{lem:cmdp_close}
If $\pi^{\zeta,*}$ is the optimal solution of (\ref{eq:cmdp_close}), then
\begin{align}
    V^{\pi^*}_{r,1}(x_1)-V^{\pi^{\zeta,*}}_{r,1}(x_1)\leq H\dfrac{\zeta}{\gamma}.
\end{align}
\end{lem}

Since the tighter optimization problem is also CMDP, we note that the second term in the right hand side of (\ref{eq:tighter_regret}) is essentially the regret of the tighter CMDP. 

Hence, from Theorem~\ref{thm:episodic} and Lemma~\ref{lem:cmdp_close} we obtain the expression of the regret bound in Theorem~\ref{thm:zero_violation}. 

\textbf{Constraint Violation}:  Again applying Theorem~\ref{thm:episodic} to the tighter optimization problem (\ref{eq:cmdp_close}), we obtain
\begin{align}
\sum_{k=1}^{K}(b+\zeta-V_{g,1}^{\pi_k}(x_1))_{+}\leq C^{\prime}\cO(\dfrac{2(1+\xi)}{\xi}\sqrt{d^3H^3T\iota^2})\nonumber
\end{align}
\begin{align}
    \sum_{k=1}^K(b+\zeta-V_{g,1}^{\pi_k}(x_1))\leq \sum_{k=1}^{K}(b+\zeta-V_{g,1}^{\pi_k}(x_1))_{+}\leq C^{\prime}\cO(\dfrac{2(1+\xi)}{\xi}\sqrt{d^3H^3T\iota^2})\nonumber
\end{align}
Hence, 
\begin{align}
    \sum_{k=1}^K(b+\zeta-V_{g,1}^{\pi_k}(x_1))\leq C^{\prime}\cO(\dfrac{2(1+\xi)}{\xi}\sqrt{d^3H^3T\iota^2})\nonumber
\end{align}
Thus,
\begin{align}
    \sum_{k=1}^K(b-V_{g,1}^{\pi_k}(x_1))\leq C^{\prime}\cO(\dfrac{2(1+\xi)}{\xi}\sqrt{d^3H^3T\iota^2})-K\zeta\nonumber
\end{align}
Hence, we have 
\begin{align}
     \sum_{k=1}^K[b-V_{g,1}^{\pi_k}(x_1)]_{+}\leq \max\{C^{\prime}\cO(\dfrac{2(1+\xi)}{\xi}\sqrt{d^3H^3T\iota^2})-K\zeta,0\}\nonumber
\end{align}
Thus, the result follows. 
\end{proof}

\textit{Proof of Lemma~\ref{lem:cmdp_close}}: We, first, introduce a few notations. 

Let $\nu^{\pi}_h(x)$ for $h=2,\ldots,H$ be
\begin{align}
    \nu^{\pi}_h(x)& =\int_{x^{\prime}}\sum_{a}\pi_h(a|x^{\prime})\phi(x^{\prime},a)^T\mu_{h-1}(x)d\nu^{\pi}_{h-1}(x^{\prime})\nonumber
\end{align}
and $\nu_1(x)$ is the initial distribution of the state. $\nu^{\pi}_h(x)$ is the distribution of the state at step $h$ while following the policy $\pi$. It is  the state occupation measure at step $h$.

Also, $\nu_h(x,a)=\pi_h(a|x)\nu_h(x)$ is the state-action occupation measure at step $h$. Hence,
\begin{align}
    V^{\pi}_{j,1}(x_1)=\sum_{h}\int_{x,a}j_h(x,a)d\nu_h(x,a)
\end{align}

Now, $\nu^{*}_h(x,a)$ corresponds to the state-action occupancy measure for the optimal policy $\pi^*$. Then, $\nu^{\zeta}_h(x,a)=(1-\zeta/\gamma)\nu^{*}_h(x,a)+\zeta/\gamma \nu^{\bar{\pi}}_h(x,a)$. Now, we are resy to prove Lemma~\ref{lem:cmdp_close}.

We have 
\begin{align}
    \sum_{h}\int_{x,a}g_h(x,a)d\nu_h^{\zeta}(x,a)\geq (1-\zeta/\gamma)b+\zeta/\gamma (b+\gamma)=b+\zeta 
\end{align}

Hence, the state-action occupancy measure $\nu^{\zeta}_1(x,a)$ is feasible for the tightened CMDP. Now, we have
\begin{align}
    \sum_{h}\int_{x,a}r_h(x,a)d\nu_h^{\zeta}(x,a)= & 
    (1-\zeta/\delta)\sum_{h}\int_{x,a}r_h(x,a)d\nu_h^*(x,a)+\zeta/\delta \sum_h\int_{x,a}r_h(x,a)d\nu_h^{\bar{\pi}}(x,a)\nonumber\\
    &\geq (1-\zeta/\delta)V_{r,1}^{*}(x_1)\nonumber
\end{align}
Since $\nu^{\zeta}_1(x,a)$ is feasible, then $V^{\pi^{\zeta,*}}_{r,1}(x_1)\geq \sum_{h}\int_{x,a}r_h(x,a)d\nu_h^{\zeta}(x,a)$. Thus, 
\begin{align}
    V_{r,1}^*(x_1)-V_{r,1}^{\zeta,*}(x_1)\leq \zeta/\delta V_{r,1}^*(x_1)\leq \dfrac{\zeta}{\delta} H
\end{align}
Hence, the result follows. \qed

\begin{remark}
Note that \cite{wei2021provably} and \cite{liu2021learning} use Lyapunov Drift analysis to obtain  constraint violation for finite state space (tabular setting). To obtain the zero violation, their approach is also similar to ours where they also consider a tighter optimization problem, and then, carefully choosing the parameter of the tighter optimization. However, one key difference is that-- they did not use any upper bound on the dual variable. Rather they rely on the Hajek's Lemma \cite{hajek1982hitting} to establish a finite bound on the dual variable for the good event. The question is whether we can use similar technique in the model-free linear function approximation. 

Note that we need to have an upper bound on the dual-variable (irrespective of the good and bad event) since the $\epsilon$-covering number (Lemma~\ref{cor:vcovering}) depends on $\xi$, the upper bound on the dual variable. Hence, we need to truncate the dual variable if it exceeds $\xi$. However, if we truncate the dual-variable, we can not use the Lyapunov-Drift analysis to bound the violation. Since in the Lyapunov-drift analysis, it relies on the fact that the magnitude of the dual-variable (or, the queue length) lower bounds the total violation (see the analysis at page 21 in \cite{liu2021learning}). Since the magnitude of the dual-variable is bounded for the good-event, one then obtain the upper bound of the violation. However, one can not extend the same analysis if we truncate the dual-variable. 

Hence, our analysis is based on the results from the convex optimization (Lemma~\ref{lem:convex}). Our approach to obtain zero violation for large enough $K$ relies on different tools compared to the Lyapunov-Drift analysis and new of a kind. 
\end{remark}

\section{Comparison with other approaches to show uniform concentration lemma  for individual value function in model-free setup}
\cite{xie2020learning} considered a zero-sum. linear Markov game setup. The paper proposed an approach where they truncate $w_h^k$, $\Lambda_h^k$ to  $\epsilon$-close value of $w$, and $\Lambda$ respectively. Subsequently, they obtained equilibrium policies using these $\epsilon$-close values. Then the proposed algorithm uses the above equilibrium policy attained using $\epsilon$-close values. Since these $\epsilon$-close values are predetermined using the $\epsilon$-covering set of $w$ and $\Lambda$ (as we have described the $\epsilon$-covering set in Lemma 14), hence, one can apply uniform concentration lemma for each individual value function \cite{xie2020learning} with error of at most $\epsilon$. We can also apply the similar trick in our set-up where we truncate the obtained $w_{j,h}^k$, and $\Lambda_h^k$ to one of the $\epsilon$-close values and then we can set the policy as the greedy one with respect to the composite approximated state-action value function.  The above method would also provide a log $\epsilon$-covering number of $\cO(\log(K))$ for each individual value function.

However, our soft-max based approach has several advantages compared to the above approach.

(i) \emph{Computation efficiency.} The alternative method explicitly rounds the $Q$-function to its $\epsilon$-close one \emph{in the algorithm}, which requires an additional $O(d^2)$ computation even with an efficient implementation (i.e., rounding on the fly)  (See Section 3.3 in \cite{xie2020learning}). In contrast, our soft-max  works directly with the actual $Q$-function without any $\epsilon$-close rounding. That is, $\epsilon$-net argument is only used \emph{in the analysis} in our setting. 

(ii) \emph{Stochastic policy.} A key fact about constrained MDPs is that the optimal policy is usually stochastic. Although the greedy-policy in the alternative method can approach the optimal policy in an average sense, in each episode, it could be far away from the optimal policy. Even though in our setting, $\alpha$ scales with $K$, our approach puts 'almost' similar probability among the composite $Q$-functions with 'almost' same values. However, in the greedy policy, it chooses the action corresponding to the highest state-action value function. Thus, such an approach can never be close to the optimal policy. 

iii) \emph{General applicability.} The alternative method relies heavily on the fact that there exists an efficient implementation of rounding on the fly in linear MDPs so that there is no need to construct an explicit $\epsilon$-net. However, beyond linear MDPs, it might not be possible to find an efficient rounding algorithm on the fly, and hence an even larger additional computation is required. In contrast, our soft-max based algorithm builds on the intrinsic smoothness-approximation trade-off in soft-max to establish uniform concentration and approximate optimism, which could potentially be generalized to other settings. Further, soft-max policy is popular, hence, our approach would provide the base for proving regret and violation bound for the general function approximation setup beyond the linear function approximation.
\section{Different episodes start from different states}
Our analysis can also be extended to the scenario when episode $k$ starts from state $x_1^k$ rather than from a fixed state. We then seek to solve the following at episode $k$
\begin{align}
\text{maximize }V_{r,1}^{\pi}(x_1^k)\quad \text{subject to }V_{g,1}^{\pi}(x_1^k)\geq b
\end{align}
$\pi^*$ is the optimal solution of the above problem.

We modify the dual update as we replace $V_{g,1}^k(x_1)$ with $V_{g,1}^k(x_1^k)$. First, it is straightforward to obtain the upper bound on $\sum_{k=1}^K(V_{r,1}^*(x_1^k)-V_{r,1}^{\pi_k}(x_1^k))+Y\sum_{k=1}^K(b-V_{g,1}^{\pi_k}(x_1^k))$ similar to Lemma~\ref{lem:dual_variable}. In particular, similar to Lemma~\ref{lem:violation}, we obtain for any $Y\in [0,\xi]$.
\begin{align}
\sum_{k}(Y-Y_k)(b-V_{g,1}^k(x_1^k))\leq \dfrac{Y^2}{2\eta}+\dfrac{\eta KH^2}{2}\nonumber
\end{align}
by replacing $V_{g,1}^k(x_1)$ with $V_{g,1}^k(x_1^k)$ at episode $k$. Then, we obtain 
\begin{align}\label{eq:decom_diff_states}
& \sum_{k=1}^K(V_{r,1}^*(x_1^k)-V_{r,1}^{\pi_k}(x_1^k))+Y\sum_{k=1}^K(b-V_{g,1}^{\pi_k}(x_1^k))\leq \dfrac{Y^2}{2\eta}+\dfrac{\eta KH^2}{2}+\nonumber\\
& \underbrace{(V_{r,1}^*(x_1^k)+Y_kV_{g,1}^*(x_1^k)-V_{r,1}^k(x_1^k)-Y_kV_{g,1}^k(x_1^k))}_{\tilde{\cT_1}}+\nonumber\\& \underbrace{(V_{r,1}^k(x_1^k)+YV_{g,1}^k(x_1^k)-V_{r,1}^{\pi_k}(x_1^k)-YV_{g,1}^{\pi_k}(x_1^k))}_{\tilde{\cT_2}}
\end{align}
We then bound $\tilde{\cT}_1$ and $\tilde{\cT}_2$ as we have done in Lemma~\ref{lem:dual_variable}.  In particular, we obtain with probability $1-p$,
\begin{align}\label{eq:diff_episode_diff_states}
\tilde{\cT}_1\leq 2H(1+\xi+H), \quad \tilde{\cT}_2\leq \cO((Y+1)\sqrt{d^3H^3T\iota^2})
\end{align}
The upper bound on regret is obtained by plugging $Y=0$.

In order to prove the constraint violation, we use Lemma 46 and Lemma  47 in \cite{ding2022provably} to obtain the following
\begin{lem}
Let Slater's condition hold, $\xi\geq 2Y^*$, and if
\begin{align}
\sum_{k=1}^K(V_{r,1}^*(x_1^k)-V_{r,1}^{\pi_k}(x_1^k))+\xi\sum_{k=1}^K(b-V_{g,1}^{\pi_k}(x_1^k))\leq \upsilon
\end{align}
then
\begin{align}
\sum_{k=1}^K(b-V_{g,1}^{\pi_k}(x_1^k))\leq \dfrac{2\upsilon}{\xi}
\end{align}
\end{lem}
It is then straightforward to bound the violation from (\ref{eq:diff_episode_diff_states}) and (\ref{eq:decom_diff_states}) with $Y=\xi$. 

In order to ensure that $\xi\geq 2Y^*$, we need to modify the Slater's condition (Assumption~\ref{assum:slater}) in the following way
\begin{assum}\label{assum:diff_states}
Let $\bar{\pi}$ be a policy such that $V_{g,1}^{\bar{\pi}}(x_1^k)\geq b+\gamma$, for $\gamma>0$.
\end{assum}
In other words, there should exist a strictly feasible policy irrespective of the initial state. Like earlier, we only need to know (or, an estimation of) $\gamma$ rather the strictly feasible policy. Then $\xi=2H/\gamma$ would ensure that $\xi\geq 2Y^*$

By combining all, we obtain that both the regret and violation are upper bounded by $\tilde{O}(\sqrt{d^3H^3T})$ and $\tilde{O}(\sqrt{d^3H^3T})$ with probability $1-p$ even when the episode starts from different states. 
\end{document}